\pdfoutput=1

\documentclass[11pt]{article}

\usepackage[preprint]{acl}

\usepackage{times}
\usepackage{latexsym}

\usepackage[T1]{fontenc}

\usepackage[utf8]{inputenc}

\usepackage{microtype}

\usepackage{inconsolata}

\usepackage{graphicx}

\usepackage{hyperref}       
\usepackage{url}            
\usepackage{booktabs}       
\usepackage{amsfonts}       
\usepackage{nicefrac}       
\usepackage{microtype}      
\usepackage{xcolor}         
\usepackage{enumitem}

\usepackage{amsmath}

\usepackage{cleveref}
\crefname{figure}{Fig.}{Figs.}
\Crefname{figure}{Fig.}{Figs.}
\crefname{table}{Tab.}{Tabs.}
\Crefname{table}{Tab.}{Tabs.}
\crefname{appendix}{App.}{Apps.}
\Crefname{appendix}{App.}{Apps.}
\crefformat{section}{\S#2#1#3} 
\crefformat{subsection}{\S#2#1#3}
\crefformat{subsubsection}{\S#2#1#3}

\usepackage{booktabs} 
\usepackage{makecell} 
\usepackage[table]{xcolor} 
\usepackage{caption} 
\usepackage{array}
\usepackage{siunitx}
\usepackage{multirow}
\usepackage{makecell}
\usepackage{tabularx}
\usepackage{wrapfig}
\usepackage{subcaption}

\usepackage{colortbl}
\definecolor{customblue}{HTML}{74AED4}
\definecolor{customgreen}{HTML}{D3E2B7}
\definecolor{customred}{HTML}{ECA8A9}
\definecolor{custompurple}{HTML}{CFAFD4}
\definecolor{customorange}{HTML}{F7C97E}
\definecolor{deepgreen}{rgb}{0.0, 0.5, 0.0}

\definecolor{citecolor}{HTML}{0071BC}
\definecolor{linkcolor}{HTML}{ED1C24}

\usepackage[most]{tcolorbox}
\usepackage{float}
\usepackage{xspace}
\tcbset{
  aibox/.style={
    width=\textwidth,
    top=10pt,
    colback=white,
    colframe=black,
    colbacktitle=black,
    enhanced,
    center,
    attach boxed title to top left={yshift=-0.1in,xshift=0.15in},
    boxed title style={boxrule=0pt,colframe=white,},
  }
}
\newtcolorbox{AIbox}[2][]{aibox,title=#2, #1, width=\textwidth, breakable}

\newtcolorbox{formattedprompt}{
  colback=green!3!white,
  colframe=green!20!white,
  fontupper=\ttfamily\small,
  breakable
}

\title{Can Multimodal LLMs See Materials Clearly? \\A Multimodal Benchmark on Materials Characterization}

\author{
    Zhengzhao Lai$^1$,
    Youbin Zheng$^2$,
    Zhenyang Cai$^1$, 
    Haonan Lyu$^3$, \\ 
    \textbf{Jingpu Yang$^2$, Hongqing Liang$^3$, Yan Hu$^{1}$\thanks{Corresponding author.}, Benyou Wang$^1$} \\ 
    $^1$The Chinese University of Hong Kong, Shenzhen \\ $^2$Northeastern University $^3$Zhejiang University \\ 
    \texttt{\{zhengzhaolai, huyan\}@cuhk.edu.cn} \\
}

\begin{document}

\maketitle

\begin{abstract}
Materials characterization is fundamental to acquiring materials information, revealing the processing-microstructure-property relationships that guide material design and optimization. While multimodal large language models (MLLMs) have recently shown promise in generative and predictive tasks within materials science, their capacity to understand real-world characterization imaging data remains underexplored. To bridge this gap, we present \textbf{MatCha}, the first benchmark for materials characterization image understanding, comprising 1,500 questions that demand expert-level domain expertise. MatCha encompasses four key stages of materials research comprising 21 distinct tasks, each designed to reflect authentic challenges faced by materials scientists. Our evaluation of state-of-the-art MLLMs on MatCha reveals a significant performance gap compared to human experts. These models exhibit degradation when addressing questions requiring higher-level expertise and sophisticated visual perception. Simple few-shot and chain-of-thought prompting struggle to alleviate these limitations. These findings highlight that existing MLLMs still exhibit limited adaptability to real-world materials characterization scenarios. We hope MatCha will facilitate future research in areas such as new material discovery and autonomous scientific agents. MatCha is available at \url{https://github.com/FreedomIntelligence/MatCha}.
\end{abstract}
\section{Introduction}
\label{section_1}

Materials characterization serves as a critical means of obtaining information about the physical world \cite{leng2013materials}, offering insights that transcend the limitations of human sensory perception. It enables multi-scale analysis, providing rich data on material morphology, composition, and structure. This information, in turn, reveals the physical, chemical, and mechanical properties essential for guiding new material design and optimization \cite{robertson2011towards}. For instance, scanning electron microscopy (SEM) and transmission electron microscopy (TEM) images are instrumental in determining the underlying mechanisms of steel bar fracture in buildings \cite{inkson2016scanning}. Despite its importance, interpreting diverse and complex imaging data generated by various characterization techniques remains a significant challenge that demands extensive domain expertise. Typically, even experienced materials researchers invest considerable time analyzing these multifaceted results. This analytical burden becomes particularly acute when handling high-throughput data, where efficiency bottlenecks arise. While convolutional neural networks (CNNs) have been utilized for various materials characterization tasks \cite{madsen2018deep, maksov2019deep, zaloga2020crystal, warmuzek2021application,leitherer2023automatic}, providing valuable insights, their application faces notable limitations. Predominantly, these CNN-based approaches are task-specific, exhibiting limited cross-task generalization. Furthermore, prior work has largely focused on morphological perception, resulting in shallow image content understanding that often falls short of the nuanced interpretations required by materials scientists in real-world scenarios. These constraints highlight the need for more versatile and deeply understanding models.

Leveraging MLLMs \cite{zhu2023minigpt, liu2023visual, li2023blip, achiam2023gpt, wang2024cogvlm, liu2024improved} for these challenges offers distinct advantages. MLLMs have demonstrated strong performance and generalization in both natural and domain-specific image understanding \cite{li2025visual}, and have spurred revolutionary changes in materials science, including property prediction \cite{rubungo2023llm, antunes2024crystal, xie2023darwin}, new material design \cite{tang2025matterchat, mishra2024foundational, xie2025darwin15largelanguage}, and autonomous scientific agents \cite{zhang2024honeycomb, ding2024matexpert}. These models show great potential to assist materials scientists with diverse materials characterization tasks via natural language interaction, thereby facilitating new material development and boosting scientific productivity. To realize this potential, MLLMs must first be capable of accurately interpreting diverse materials characterization images, recognizing fundamental visual content, and performing reasoning based on visual cues. However, existing evaluations of MLLMs on scientific imaging data primarily focus on biomedical domains \cite{he2020pathvqa, huang2023visual, NEURIPS2024_36b31e1b}, are confined to relatively simplistic figures and charts in limited scientific fields \cite{yue2024mmmu, yue2024mmmu-pro, chen2024we, NEURIPS2024_217bb44a}, or lacking sufficient depth to represent the complexity of authentic materials research scenarios \cite{alampara2024macbench, li2025mmsci, verma2024beyond}. Consequently, a comprehensive, expert knowledge-anchored multimodal benchmark specifically designed for materials science to rigorously assess current models is notably absent. This gap impedes the progress toward AI-assisted research and autonomous scientific discovery agents.

To bridge this gap, we present MatCha, a challenging multimodal benchmark for materials characterization imaging data understanding. 
The core strengths of MatCha are threefold: (1) \textit{practical and realistic task design}: The design philosophy of MatCha originates from real-world scientific workflows. The tasks are derived directly from the research processes of materials scientists and are designed to reflect authentic challenges in practice. (2) \textit{task diversity and broad coverage}: MatCha incorporates 21 sub-tasks, each representing a concrete step within the scientific workflow. These tasks collectively cover a wide range of characterization methods and corresponding problems. (3) \textit{expert-level difficulty}: MatCha includes 1,500 multiple-choice questions of varying complexity, each requiring visual understanding and expert-level scientific expertise.

We first benchmark state-of-the-art MLLMs on MatCha under the zero-shot setting, observing a substantial performance gap between models and human experts as well as noticeable performance degradation across different task stages due to limited generalization capability. Next, we further investigate whether these performance gaps can be bridged by in-context learning or by guiding the model through a chain-of-thought (CoT) process. The results show that while some models do benefit from these strategies, others exhibit unstable or even degraded performance, and a significant gap to human expert performance persists.

In summary, our contributions are as follows:
\begin{itemize}
  \vspace{-2mm}
  \item We introduce MatCha, the first multimodal perception and understanding benchmark for materials characterization, which comprises 21 expert-defined tasks reflecting real scientific challenges.
  \vspace{-2mm}
  \item We conduct extensive experiments on various proprietary and open-source models under different settings.
  \vspace{-2mm}
  \item We reveal their current limitations and directions for future enhancements through detailed quantitative and qualitative analyses.
\end{itemize}
\section{Related Works}
\label{section_2}

\paragraph{Materials Characterization Analysis.}

Computer vision has revolutionized the extraction of visual information and quantitative analysis from material microscopy images. Applications include image-based structure recognition \cite{abouelatta2013classification, decost2015computer, chowdhury2016image, decost2017exploring}, detection of individual atomic sites and defects \cite{madsen2018deep,li2018automated,  yang2021deep, shen2021multi}, and segmentation of microstructures or particles \cite{decost2019high, roberts2019deep, baskaran2020adaptive, bals2023deep}. However, existing studies largely focus on perception in specific electron microscopes, neglecting deeper imaging content and cross-modal data analysis, which results in a gap with actual scientific research problems. Additionally, material science characterization data significantly differ from natural images, thus limiting generalization between tasks. Furthermore, the scarcity of large labeled datasets impedes supervised deep learning model training \cite{holm2020overview}. Our research seeks to leverage and investigate the capabilities of MLLMs to address the diverse challenges in characterization data analysis and elucidation, thereby tackling real-world problems faced by materials scientists.

\paragraph{Multimodal Benchmarks in Science.}

Recent advancements in MLLMs \cite{liu2023visual, zhu2023minigpt, li2023blip, NEURIPS2023_9a6a435e} have spurred the development of benchmarks to evaluate their scientific problem-solving capabilities. For example, MMMU \cite{yue2024mmmu} and MMMU-Pro \cite{yue2024mmmu-pro} present extensive multi-discipline college-level tasks. In materials and chemistry, MaScQA \cite{zaki2024mascqa} offers a text-based question dataset, while MaCBench \cite{alampara2024macbench} proposes a multimodal benchmark. However, MaCBench \cite{alampara2024macbench} concentrates on chemistry and general laboratory knowledge, with limited content on crystalline materials. SciFIBench \cite{NEURIPS2024_217bb44a} targets scientific figure interpretation using arXiv papers, which do not cover materials science or chemistry extensively \cite{hsu-etal-2021-scicap-generating, li-etal-2024-multimodal-arxiv} and, being non-peer-reviewed, may have quality concerns. While MMSci \cite{li2025mmsci} sources its content from Nature Communications, its task formats--captioning and figure-caption matching--are not representative of the queries posed in actual scientific research, which limits its practical applicability. MatCha, in contrast, is designed with problem formats that authentically mirror the challenges scientists face during the process of scientific discovery.
\section{The MatCha benchmark}
\label{section_3}

\begin{figure*}[!t]
  \centering
  \includegraphics[width=1\textwidth, height=13cm, keepaspectratio]{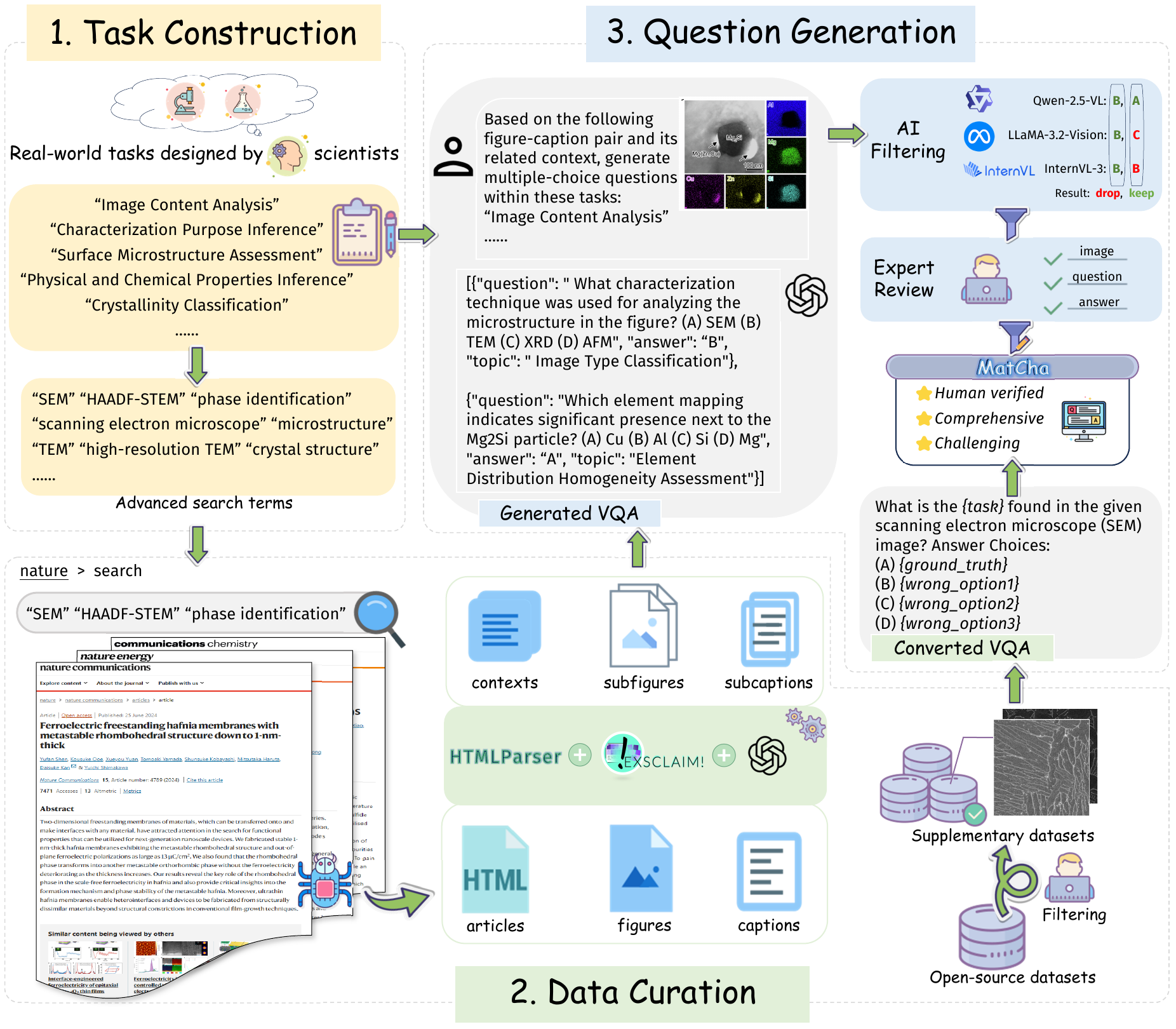}
  \caption{MatCha construction pipeline. First, experts define scientifically meaningful, practical tasks and extract key terms. Second, data is collected and processed using these terms. Third, GPT-based generation and template-based conversion are employed to construct samples from the gathered data, followed by quality filtering and review.}
  \vspace{-\intextsep}
  \label{fig:pipeline}
\end{figure*}

We introduce MatCha, a comprehensive and challenging benchmark for advancing multimodal materials characterization visual analysis and understanding. MatCha comprises 1,500 questions across 21 tasks, reflecting expert-level difficulty and grounded in real-world scientific scenarios faced by scientists. In the following sections, we elaborate on how we construct tasks (\cref{3.1}), collect data (\cref{3.2}), generate questions (\cref{3.3}), and analyze its composition (\cref{3.4}). The holistic construction pipeline is shown in \cref{fig:pipeline}.

\subsection{Task Construction}
\label{3.1}

To cover a wide range of domain-specific knowledge for a comprehensive evaluation, we collaborate with experienced researchers in materials science. Drawing from the typical workflow of materials science research--"Processing" → "Morphology" → "Structure" → "Property"--we designed a corresponding chain of stages that reflects this logical progression. The details are as follows:

\paragraph{Processing Correlation (PC).}
As a foundational step in materials characterization, it involves identifying the characterization technique and its intended purpose. It evaluates the ability to accurately awareness characterization methods and their appropriate application contexts.

\paragraph{Morphology Analysis (MA).}
This stage focuses on observing and assessing the surface or cross-sectional of materials to gather morphological information. It measures the capability of models to perceive both macro- and micro-scale visual characteristics in electron microscopy images.

\paragraph{Structure Analysis (SA).}
It targets the interpretation of material structure at the micro- or atomic-scale, which is essential for understanding the underlying mechanisms of various properties, such as mechanical behavior. It further assesses the ability to integrate and link cross-modal knowledge, for instance, linking spectral peaks to chemical bonds or functional groups.

\paragraph{Property Analysis (PA).}
Since the structure of a material determines its properties, this stage poses a more difficult challenge, by evaluating the logical reasoning capabilities in connecting structural features with properties in physical and chemical.

To concretize these research stages, we define several scientifically meaningful sub-tasks for each, as detailed in \cref{sec:appendix_subtask}. These stages represent the essential steps that materials scientists typically undertake, while the sub-tasks capture a diverse range of real-world challenges encountered during new materials development. Evaluating MLLMs on these tasks is crucial to reveal their potential and limitations in scientific research, providing critical insights into their capability boundaries and guiding future advancement for applications.

\subsection{Data Curation}
\label{3.2}

\paragraph{Data Collection.}
Given the diversity of characterization data, firstly, we define a set of advanced search terms and their synonyms that exhaustively cover the types of data required across the various sub-tasks. With these terms, we employ Exsclaim \cite{schwenker2021exsclaim} to search for and retrieve publicly accessible articles from the Nature platform under CC BY-4.0 license. The search results are sorted by relevance. For each article, we download the HTML file along with all associated figures and their corresponding captions. We crawl 340 articles containing 2,165 figures in total.

\paragraph{Data Processing.}
First, we query GPT-4o with prompts presented in \cref{sec:appd_prompt_1} to segment the full caption of each figure into sub-captions corresponding to sub-figures. Second, Exsclaim \cite{schwenker2021exsclaim} is applied to split each figure into its constituent sub-figures and assign the appropriate sub-captions to each one. The resulting sub-figures are categorized. To ensure the authenticity of the dataset, we retain categories that accurately reflect real-world materials characterization scenarios and exclude other simulated types of sub-figures, such as illustrations. To compensate for insufficient information in the subsequent VQA generation caused by the overly short sub-captions, we use a parser from \cite{toland.2023.accelerated.scheme} to parse HTML files of the articles and extract the main body text. We then develop a regular expression matching function to retrieve the context relevant to each sub-figure from the main content. Finally, these remaining article contents and images form the basis for the subsequent question generation.

\paragraph{Data Supplementation.}
Figures in published papers often inevitably contain annotations or markings that may hint at microstructural characteristics. Moreover, although we download high-resolution figures directly, their visual clarity can still be inferior to images captured directly from characterization instruments. To alleviate this and enhance benchmark diversity and challenge, we source additional non-simulated, human-annotated datasets. Following rigorous expert review and filtering, three datasets \cite{uhcsdata, baskaran2020adaptive, dennler2021learning} are selected as supplementary data sources. These datasets consist of high-quality, real-world electron microscopy images, and are used to construct supplementary sub-tasks: surface microstructure analysis (Suppl. SMA), defect type classification (Suppl. DTC), and image content analysis (Suppl. ICA). Specifically, Suppl. SMA requires analyzing the microstructural features of Ti-6Al-4V alloy images; Suppl. DTC examines defect identification and distinguishment between different defect types; and Suppl. ICA assesses the analysis of primary microstructural components in low-carbon steel. These supplementary tasks focus on common microstructural analyses in practical yet fundamental materials scenarios, evaluating the domain-specific knowledge and visual perception capabilities of MLLMs.

\begin{figure*}[t]
  \centering

  \begin{minipage}[t]{0.30\textwidth}
    \centering
\scalebox{0.51}{
  \begin{tabular}{lr}
  \hline
  \toprule
  \textbf{Statistics}            & \textbf{Value}   \\
  \midrule

  \textbf{MatCha Instances}      &                  \\
  Total Images                   & $1,260$          \\
  Total QA samples               & $1,500$          \\
  Average size (px)              & $559\times660$   \\
  Maximum size (px)              & $1876\times1064$ \\
  \midrule

  \textbf{Processing Correlation} &                  \\
  - questions                    & $153$          \\
  - average length               & $160$           \\
  - options                    & ${4^{\prime}: 153}$          \\
  \midrule

  \textbf{Morphology Analysis}   &                  \\
  - questions                    & $795$          \\
  - average length               & $236$           \\
  - \makecell[l]{options \\ $ $}                    & \makecell[r]{$7^{\prime}: 206, 5^{\prime}: 4, $ \\ $4^{\prime}: 308, 3^{\prime}: 275, 2^{\prime}: 2$}          \\
  - \# Among them,              &                  \\
  for the supplementary             &                  \\
  - questions                    & $506$          \\
  - average length               & $271$           \\
  - options                    & ${7^{\prime}: 206, 4^{\prime}: 33, 3^{\prime}: 267}$          \\

  \textbf{Structure Analysis} &                  \\
  - questions                    & $370$          \\
  - average length               & $187$           \\
  - \makecell[l]{options \\ $ $}                    & \makecell[r]{$6^{\prime}: 1, 5^{\prime}: 6, $ \\ $4^{\prime}: 345, 3^{\prime}: 15, 2^{\prime}: 3$}          \\

  \textbf{Property Analysis} &                  \\
  - questions                    & $182$          \\
  - average length               & $194$           \\
  - options                    & ${5^{\prime}: 4, 4^{\prime}: 163, 3^{\prime}: 12, 2^{\prime}: 3}$          \\

  \midrule

  \bottomrule
  \hline
  \end{tabular}
  \vspace{-\intextsep}
  }
    \caption{Statistical overview of MatCha samples. Superscripts indicate the number of multiple-choice options for those questions.}
    \label{tab:statistics}
  \end{minipage}
  \hfill
  \begin{minipage}[t]{0.68\textwidth}
    \centering
    \vspace{-17.2ex}
    \includegraphics[width=0.9\linewidth]{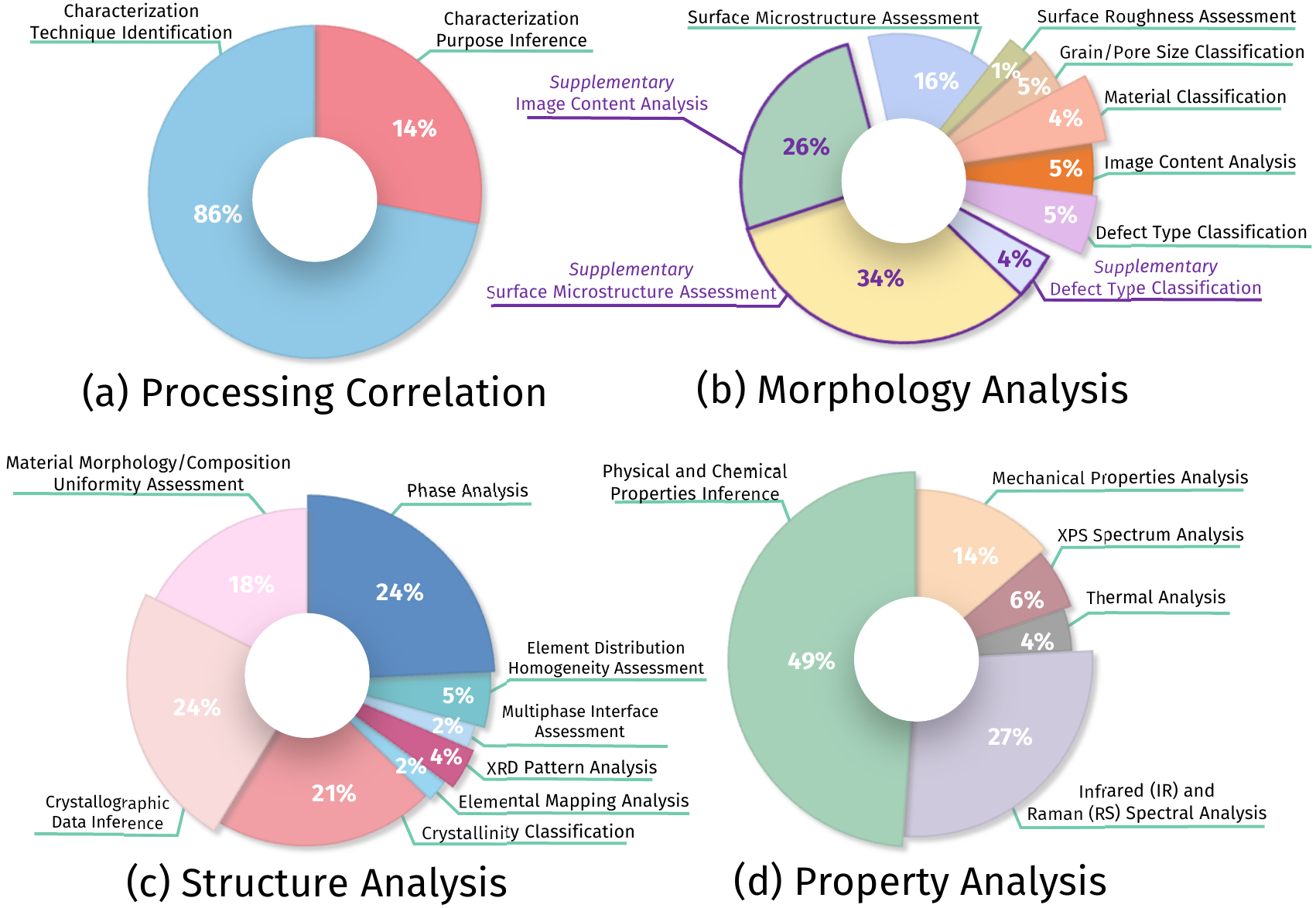}
    \caption{Composition of the MatCha benchmark, illustrating the proportion and distribution statistics of its 21 sub-tasks across the four progressive research stages: Processing Correlation (PC), Morphology Analysis (MA), Structure Analysis (SA), and Property Analysis (PA).}
    \label{fig:statistics}
  \end{minipage}
   \vspace{-\intextsep}
\end{figure*}

\subsection{Question Generation}
\label{3.3}

We formulate our benchmark in a closed-ended VQA format. This approach facilitates easier analysis compared to open-ended VQA and obviates the need for LLMs in automatic evaluation. Consequently, it eliminates subjectivity in answer assessment, ensuring that evaluation results more accurately reflect the true performance of MLLMs.

\paragraph{Generated VQA.}
Equipped with collected (sub-figure, sub-caption, context) triplets, we employ GPT-4o to generate multiple-choice questions. Considering the unique nature of images in each article and to prevent overly image-specific or divergent questions lacking generalizability, we carefully design a prompt that constrains VQA generation within our predefined sub-task scopes. However, we do not restrict the number of answer choices, allowing the model to fully utilize the unique information in each data triplet, as detailed in \cref{sec:appd_prompt_2}. Each generated VQA sample includes a sub-figure, a question within a sub-task, and a set of answer choices, where one is correct and the others are distractors generated by GPT-4o. This process finally yields 26,891 samples, which subsequently undergo data filtering and expert review.

\paragraph{Converted VQA.}
For the three supplementary datasets, we convert their metadata and labels into multiple-choice questions using a QA template. Similarly, each converted VQA sample consists of an image, a question designed under a certain sub-task, and several answer choices. One choice is the correct answer derived from the original label, while the remaining options are distractors sampled from the label set, excluding the correct one.

\paragraph{Data filtering.}
For the generated VQA samples, we perform a coarse-grained filtering using AI experts as the first step. Concretely, we use Qwen2.5-VL-7B \cite{Qwen2VL}, InternVL3-8B \cite{chen2024internvl}, and LLaMA-3.2-11B-Vision \cite{grattafiori2024llama} to each attempt the question three times. If all AI experts answer the question correctly in all attempts, the question is deemed too simple and removed. The remaining questions are retained. This process effectively filters out easy questions and preserves those both challenging and discriminative, ensuring quality.

\paragraph{Experts review.}
Following coarse filtering by AI experts, two materials science experts--Ph.D. candidates from the material science and engineering department--conduct a manual review to ensure: 1) each figure is a real photograph or data-generated plot, not a simulation or illustration, ensuring data authenticity; 2) each question is grounded in visual content, answerable solely through visual cues and reasoning with intrinsic domain knowledge, without requiring external contextual information, ensuring question validity; 3) overly simple optical character recognition (OCR)-style questions lacking domain-specific expertise are removed to emphasize professionalism and challenge, reflecting realistic scenarios for materials researchers. After this, 994 samples remain, forming the core of MatCha. Additionally, as the three supplementary datasets have already undergone manual validation and annotation, we randomly select 506 samples from their converted VQA set. Together, these comprise a total of 1,500 samples that constitute the complete benchmark. In \cref{sec:appendix_vqa_cases}, we provide some examples as illustrations.

\subsection{Analysis of MatCha Benchmark}
\label{3.4}

Scientific discovery in materials science demands multiple and complex characterizations beyond basic natural image perception and shallow domain expertise. To investigate the diversity and representativeness of our benchmark, we count and analyze the distribution of sub-tasks, characterization techniques, and material types. A detailed statistical breakdown is provided in \cref{sec:appendix_data_stats}. The quantitative statistics show that the dataset encompasses a wide array of methods. The material types include metallic materials, inorganic non-metallic materials, composites, and organic polymers. Furthermore, the collected source articles are retrieved from 14 different journals under the Nature platform, with publication dates ranging from 2015 to 2025, mitigating potential biases towards specific research topics or time periods. This broad scope ensures MatCha reflects a wide range of real-world scientific challenges. The statistics of different sub-task samples are shown in \cref{tab:statistics} and \cref{fig:statistics}.
\section{Experiments}
\label{section_4}

\subsection{Experimental Setup}

\definecolor{lightblue}{RGB}{222,235,247}
\definecolor{lightyellow}{RGB}{255,242,204}
\definecolor{lightred}{RGB}{249,219,219}

\begin{table*}[t]
    \resizebox{\textwidth}{!}{
    \begin{tabular}{l|cccccccccc}

    \hline
    \toprule
    \multirow{2}{*}{\textbf{Model}} & \multicolumn{5}{c}{{\cellcolor{lightyellow}\textbf{Generated VQA}}} & \multicolumn{4}{c}{{\cellcolor{lightblue}\textbf{Converted VQA}}} & \multicolumn{1}{c}{\cellcolor{lightred}\textbf{MatCha}}\\

    & \cellcolor{lightyellow}\textbf{PC} & \cellcolor{lightyellow}\textbf{MA} & \cellcolor{lightyellow}\textbf{SA} & \cellcolor{lightyellow}\textbf{PA} & \cellcolor{lightyellow}\textbf{All} & \cellcolor{lightblue}\textbf{Suppl. SMA} & \cellcolor{lightblue}\textbf{Suppl. DTC} & \cellcolor{lightblue}\textbf{Suppl. ICA} & \cellcolor{lightblue}\textbf{All} & \cellcolor{lightred}\textbf{All} \\

    \midrule
        \multicolumn{11}{c}{\cellcolor[HTML]{F5F5F5}\textit{Baselines}} \\
    \midrule

    Random Choice               & 19.61  & 26.64  & 23.24   & 24.72  & 15.79  & 34.83  & 24.24 & 15.53 & 23.91 & 24.73\\
    Human                       & 90.26  & 89.31  & 87.57  & 88.59  & 88.87  & 94.76  & 87.88  & 81.55 & 88.93 & 88.89 \\
    
    \midrule
        \multicolumn{11}{c}{\cellcolor[HTML]{F5F5F5}\textit{Proprietary Models}} \\
    \midrule
    
    GPT-4o \cite{achiam2023gpt}                      & \underline{85.62} & \underline{70.59}  & \textbf{71.62} & \underline{66.48}  & \textbf{62.58} & \underline{60.30} & 63.64 & 39.81 & \underline{52.17} & \textbf{59.07} \\
    GPT-4o-mini \cite{hurst2024gpt}                 & 66.67  & 61.59 & 57.57  & 51.65 & 45.27 & 54.31 & 48.48 & 18.93 & 39.53 & 43.33 \\
    Gemini-1.5-Flash \cite{team2024gemini}            & 75.16 & 62.63 & 56.76 & 54.95 & 48.39 & 46.82 & 48.48 & 31.07 & 40.51 & 45.73 \\
    Gemini-1.5-Pro \cite{team2024gemini}              & \textbf{86.27} & 70.24 & \underline{67.03} & 62.64 & 59.76 & 55.43 & \underline{66.67} & 39.81 & 49.80 & 56.40 \\
    Claude-3.5-Sonnet \cite{claude}           & 83.66 & \textbf{71.28} & 66.49 & \textbf{68.68} & \underline{61.37} & 50.56 & \textbf{75.76} & \textbf{46.60} & 50.59 & \underline{57.73} \\
    LlaMA-4-Maverick \cite{meta2025llama}            & 79.74 & 62.98 & 60.81 & 59.89 & 53.32 & \textbf{69.66} & 48.48 & \underline{43.69} & \textbf{57.71} & 54.80 \\

    \midrule
        \multicolumn{11}{c}{\cellcolor[HTML]{F5F5F5}\textit{Open-source Models}} \\
    \midrule
    
    Qwen2.5-VL-7B \cite{qwen2_5}               & 66.01 & \underline{57.44} & 54.05 & 53.85 & 43.46 & 45.69 & \textbf{54.55} & 20.87 & 36.17 & 41.00 \\
    Qwen2.5-VL-32B \cite{qwen2_5}             & \textbf{69.28} & \textbf{66.44} & \textbf{60.81} & \textbf{66.48} & \textbf{52.62} & \underline{58.43} & \textbf{54.55} & \underline{25.24} & \underline{44.66} & \textbf{49.93} \\
    InternVL3-8B \cite{chen2024expanding}                & 41.83 & 50.87 & 48.38 & 48.90 & 36.12 & 46.82 & \textbf{54.55} & 21.36 & 36.96 & 36.40 \\
    InternVL3-38B \cite{chen2024expanding}               & \underline{67.97} & \textbf{66.44} & \underline{58.38} & \underline{60.99} & \underline{49.70} & \textbf{63.30} & \textbf{54.55} & 23.79 & \textbf{46.64} & \underline{48.67} \\
    LLaVA-1.5-7B \cite{liu2024improved}                & 22.22 & 26.99 & 20.00 & 29.12 & 15.90 & 41.57 & 24.24 & 14.56 & 29.45 & 20.47 \\
    LLaVA-1.5-13B \cite{liu2024improved}               & 43.14 & 40.14 & 38.92 & 37.36 & 29.18 & 33.71 & 27.27 & 14.56 & 25.49 & 27.93 \\
    Llama-3.2-11B-Vision \cite{grattafiori2024llama}        & 60.13 & 40.48 & 40.27 & 41.21 & 31.39 & 38.95 & 12.12 & 18.45 & 28.85 & 30.53 \\
    Janus-Pro-7B \cite{chen2025janus}                & 48.37 & 49.13 & 51.08 & 53.85 & 39.54 & 25.09 & \underline{48.48} & 19.90 & 24.51 & 34.47 \\
    Gemma-3-4b-it \cite{team2025gemma}               & 60.13 & 47.06 & 43.24 & 41.21 & 34.41 & 39.33 & 45.45 & \textbf{25.73} & 34.19 & 34.33 \\

    \bottomrule
    \hline
    \end{tabular}
    }
    \caption{Evaluation results of model performance on MatCha.
     Generated VQA: PC (processing correlation), MA (morphology analysis), SA (structure analysis), PA (property analysis). Converted VQA: Suppl. SMA (supplementary surface microstructure analysis), Suppl. ICA (supplementary image content analysis), Suppl. DTC (supplementary defect type classification).
    \textbf{Bolded} values signify the optimal in-class outcomes (open-source or proprietary) and \underline{underlined} values indicate the suboptimal performance.
    }
    \label{tab:main_results}
    \vspace{-\intextsep}
\end{table*}

\paragraph{Evaluation models.}
We use random selection as a baseline, where a randomly chosen option is treated as the answer. We evaluate a diverse set of MLLMs, including the following proprietary models: GPT-\{4o \cite{achiam2023gpt}, 4o-mini \cite{hurst2024gpt}\}, Gemini-1.5-\{Pro \cite{team2024gemini}, Flash \cite{team2024gemini}\}, Claude-3.5-Sonnet \cite{claude}, Llama-4-Maverick \cite{meta2025llama}. We also evaluate the following popular open-source models: LLaVa-1.5 series \cite{liu2024improved}, Qwen-2.5-VL series \cite{qwen2_5}, InternVL-3 series \cite{chen2024expanding}, Llama-3.2-Vision \cite{grattafiori2024llama}, Janus-Pro \cite{chen2025janus}, and Gemma-3 \cite{team2025gemma}. Notably, we choose their chat or instruction-tuned version for each model for better capability of instruction following.

\paragraph{Implementation details.}
Since all questions are in multiple-choice format, we instruct the models to constrain their responses to the provided option letters. A detailed prompt can be found in \cref{sec:appd_prompt_3}. For proprietary models, inference is conducted via the API platform. For open-source models, we utilize the Transformers library \cite{wolf2020transformers} and the LLaMA-Factory toolkit \cite{zheng2024llamafactory} to perform inference on NVIDIA RTX 6000 GPUs. More details are presented in \cref{sec:appendix_hyp}.

To better understand the performance gap between models and humans, we invite doctoral researchers familiar with various characterization techniques from the material science and engineering department to participate in tests as a human baseline. They are presented with the same format of questions and instructions as the models to ensure a fair comparison.

\subsection{Experimental Results}

We adopt a generic zero-shot strategy for evaluation, and the quantitative results of 6 proprietary models and 9 open-source models with varying sizes and architectures are in \cref{tab:main_results}. Considering the different sources of data, we report the Generated VQA subset and the Converted VQA subset, respectively. The main results are as follows:

\paragraph{MatCha represents a challenging benchmark.}
On the generated VQA subset, the best-performing model, GPT-4o, achieves only 62.58\% accuracy, exhibiting a 26.29\% gap from human performance (88.87\%). When facing the more difficult converted VQA subset, the top-performing model, LLaMA-4-Maverick, scores only 57.71\%, significantly lower than the 88.93\% accuracy attained by human experts. These results demonstrate that MatCha provides a well-calibrated level of difficulty and highlight the considerable performance gap between current MLLMs and human experts in materials characterization. 

A no-image ablation study is also conducted in \cref{sec:appendix_abla} to validate that the questions in MatCha are strongly grounded in visual content.

\paragraph{Proprietary models outperform open-source models.}
Considering the overall tasks in the generated VQA subset, the leading proprietary model outperforms its open-source counterpart by 9.96\% (62.58\% to 52.62\%), indicating a considerable performance gap. This disparity widens to 11.07\% (57.71\% to 46.64\%) in the converted VQA subset. Overall, open-source models explicitly exhibit low performance across both generated and converted VQA subsets, with most models failing to correctly answer more than 40\% of the questions.

\paragraph{Performance and generalization capability disparities exist across models.}
Models show considerable differences in performance across task dimensions, and even within the same model, there are marked fluctuations across different research stages (\emph{e.g.}, PC vs. PA). These results suggest that current models struggle with task generalization and knowledge transfer within the materials science domain.

\subsection{Analysis}

\paragraph{Most models still struggle with relatively simple perceptual tasks.}
Some proprietary models, such as Gemini-1.5-Pro, GPT-4o, and Claude-3.5-Sonnet, lead in tasks from the PC and MA stages. Gemini-1.5-Pro, for instance, trails human performance by only 3.99\% on the PC stage, indicating its reasonable ability to handle basic characterization techniques. However, even the best-performing model on the MA stage--Claude-3.5-Sonnet--still lags behind human experts by approximately 18.03\%. This highlights that, despite recent advancements in understanding natural images, current cutting-edge MLLMs remain insufficient for achieving expert-level pattern recognition in fine-grained morphological analysis of materials imaging. We attribute this to the current lack of high-quality scientific training corpora, which hinders models from generalizing the ability developed on natural images to these specialized tasks.

Nevertheless, beyond these three models, the performance of most other proprietary and open-source models remains suboptimal. For example, most open-source models failed to exceed 50\% accuracy for tasks in the MA stage. This suggests that many current open-source models still struggle to extract and interpret morphological features from electron microscopy images, particularly when recognizing multiscale structures. However, Qwen2.5-VL-32B and InternVL3-38B achieve a notable 66.44\% on the MA stage, outperforming certain proprietary models such as GPT-4o-mini (61.91\%). This indicates the potential for open-source models to match or even surpass general-purpose proprietary models in specialized scientific domains when properly optimized.

\begin{figure*}[!t]
  \centering
  \includegraphics[width=1\textwidth]{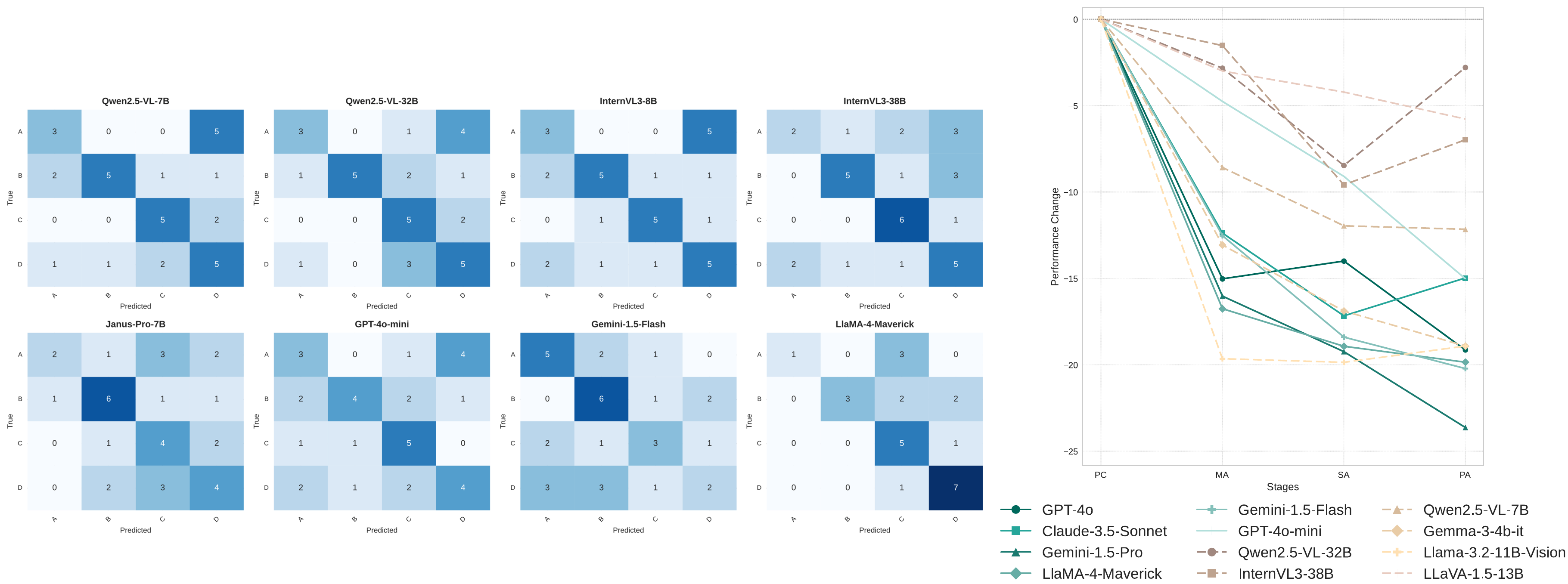}
  \caption{Performance analysis. Left: Confusion matrices of several models on the Suppl. DTC task. Right: Performance degradation trends of models across progressive stages.}
  \label{fig:combined}
   \vspace{-\intextsep}
\end{figure*}

\paragraph{Most models degrade on tasks requiring more expertise and reasoning ability.}
Benefiting from our stage-based task design, we can conduct a fine-grained analysis of both proprietary and open-source models across different materials research stages, as this design aligns well with the research workflow of materials scientists. As shown in the right figure of \cref{fig:combined}, we observe a clear trend: as tasks progress from morphology analysis to structure interpretation and properties elucidation--stages requiring deeper materials domain knowledge and reasoning ability--most models exhibit a significant decline in performance. Interestingly, this performance deterioration is more pronounced among proprietary MLLMs. On average, proprietary models show a 15.96\% accuracy degradation across the MA, SA, and PA stages. In contrast, open-source models demonstrate a smaller average decline of 10.29\%. Consequently, several open-source models, including Qwen2.5-VL-7B, Qwen2.5-VL-32B, and InternVL3-38B, have outperformed certain proprietary models in the MA, SA, and PA stages, in some cases approaching the accuracy of the best proprietary models. We hypothesize that this may be related to the incorporation of a larger proportion of scientific imagery or domain-specific textual data during training, particularly in scientific discourse.

These findings suggest that current mainstream MLLMs still lack sufficient materials science domain-specific understanding and multi-level reasoning capabilities, rendering them inadequate for research tasks that demand deep materials expertise and complex logical analysis. This limitation constrains their potential for broader application in scientific research support.

\paragraph{Models fail to recognize microstructural details in more realistic and complex perceptual scenarios.}
In the converted VQA subset, which involves real-world materials electron microscopy characterization images, current MLLMs encounter substantial challenges despite these tasks being relatively basic for human experts. In the Suppl. SMA task, LLaMA-4-Maverick achieves the best performance with an accuracy of 69.66\%, still approximately 25.1\% lower than human performance. Open-source models struggle to identify the microstructural hierarchy in alloy images, with most accuracies falling between 30\% and 50\%. In the Suppl. DTC task, which requires the detection and classification of subtle structural defects, both proprietary and open-source models exhibited significant difficulties. Many failed to determine the presence of defects or distinguish between defects from different origins, indicating a breakdown in their recognition capabilities, as shown in the left figure of \cref{fig:combined}. The Suppl. ICA task poses an even greater challenge, demanding fine-grained differentiation of multiple microstructural components in low-carbon steel. This task places high demands on both visual discrimination and metallography knowledge. All open-source models achieve accuracies below 30\%, and proprietary models also perform suboptimally, revealing severe limitations in the understanding and reasoning abilities of current models regarding materials microstructure.

Notably, GPT-4o consistently ranked among the top three performers across all three supplementary tasks, suggesting a relatively stronger generalization capability. This may be attributed to its more advanced multimodal alignment mechanisms and broader pretraining data coverage.

\subsection{Impact of Few-shot and Chain-of-Thought}

To investigate whether the performance limitations observed in the zero-shot setting can be mitigated, we conduct further experiments using few-shot in-context learning and CoT prompting. The detailed results of performance across various few-shot settings and CoT are presented in \cref{sec:appendix_fewshot_cot}.

\paragraph{Few-shot Learning}
Our few-shot experiments reveal that the ability to leverage in-context examples varies dramatically across models. On the one hand, some models show significant positive scaling. GPT-4o, for instance, demonstrates a remarkable improvement on the challenging Converted VQA subset, with its accuracy jumping from 52.17\% (zero-shot) to 73.52\% (16-shot). Other models like Claude-3.5-Sonnet and GPT-4o-mini also show a clear upward trend. This suggests that for some models, in-context examples can effectively elicit domain-specific knowledge and improve pattern recognition. On the other hand, this benefit is not universal and can even be detrimental. For Gemini-1.5-Pro, performance is inconsistent and can even degrade at higher shot counts. The effect is more severe for Qwen2.5-VL-32B, which suffers a notable performance drop when provided with examples. These cases indicate that for certain models, in-context examples may introduce noise or conflict with their internal knowledge, making the effective utilization of such prompts a non-trivial challenge.

\paragraph{Chain-of-Thought Prompting}

Our evaluation of CoT prompting reveals mixed results. On one hand, most models show slight improvement with step-by-step reasoning. Janus-Pro, for instance, sees the most significant benefit, with its accuracy increasing by nearly 6 percentage points to 40.27\%. On the other hand, for top-performing models like Gemini-1.5-Pro and Llama-4-Maverick, CoT prompting actually hindered performance compared to the direct zero-shot approach.

These contradictory outcomes suggest that CoT is not a universally effective strategy. While it can help some models, it may conflict with the internal reasoning pathways of others. This reinforces our finding that simple prompting techniques, such as in-context learning and CoT, are insufficient to overcome fundamental gaps in intrinsic domain knowledge and visual perception. Therefore, other techniques, such as the retrieval mentioned in \cref{sec:appendix_rag}, remain for future exploration.

\subsection{Error Analysis}

To gain deeper insights into the nature of model failures, we sample 100 incorrect predictions from GPT-4o and categorize the primary cause of error using GPT-4o. We predefine four main error types: 1) \textit{Visual Perception Error}, where the model fails to correctly identify visual features, such as structures, boundaries, or grains; 2) \textit{Lack of Material Knowledge}, where the model misunderstands or lacks necessary materials domain concepts; 3) \textit{Language and Logic Failure}, involving errors in explanation or the semantic understanding of the question and options; and 4) \textit{Image-Text Alignment Misunderstanding}, where the model incorrectly links image content to the question or options text.

\begin{table}[t]
    \centering

    \setlength{\tabcolsep}{6pt} 

    \caption{Distribution of error types for three models.}
    \label{tab:error_analysis}
    \resizebox{\columnwidth}{!}{
    \begin{tabular}{lcccc}
        \toprule

        \textbf{Model} & \makecell{Visual \\ Perception \\ Error} & \makecell{Lack of \\ Material \\ Knowledge} & \makecell{Language \& \\ Logic \\ Failure} & \makecell{Image-Text \\ Alignment \\ Error} \\
        \midrule
        GPT-4o \cite{achiam2023gpt} & 27\% & 71\% & 1\% & 1\% \\
        GPT-4o-mini \cite{hurst2024gpt} & 34\% & 59\% & 5\% & 2\% \\
        Gemini-1.5-Pro \cite{team2024gemini} & 32\% & 64\% & 2\% & 2\% \\
        \bottomrule
    \end{tabular}
    }
    \vspace{-\intextsep}
\end{table}

As shown in \cref{tab:error_analysis}, the predominant error category across all models is \textit{Lack of Material Knowledge}, accounting for over 60-70\% of failures, confirming that a core limitation of current MLLMs is their deficiency in specialized scientific domain knowledge. \textit{Visual Perception Error} is the second most common failure type, highlighting that even leading models struggle with the fine-grained and complex visual patterns in scientific imagery, a conclusion that aligns with our previous analysis. We provide error cases for illustration in \cref{sec:appendix_error_cases}.
\section{Conclusion}
\label{section_5}

Automated interpretation of complex materials characterization imaging data remains a significant bottleneck. In this work, we introduce MatCha, the first benchmark designed to assess the capability of current MLLMs to understand materials characterization imagery. We construct a suite of realistic and scientifically meaningful tasks to evaluate state-of-the-art MLLMs. Our comprehensive evaluations pinpoint their inadequacy for generalization, deep domain knowledge, complex morphology perception, and nuanced materials analysis. MatCha thus serves as a critical tool for diagnosing these core deficiencies, aiming to guide the development of MLLMs that can truly accelerate materials research and enable autonomous scientific discovery.
\section*{Limitations}

Although MatCha aims for broad coverage with 21 sub-tasks, the vast field of materials science means it cannot encompass every material, characterization technique, experimental nuance, or specific research question. Certain emerging areas or highly specialized analyses might be underrepresented. Meanwhile, the dataset of 1,500 questions, while reviewed by experts, might inadvertently contain biases or not fully capture the statistical distribution of all real-world scenarios.

\section*{Potential Risks}

While MatCha is designed to rigorously assess the specialized knowledge and visual understanding capabilities of MLLMs within the domain of materials characterization, a potential risk or limitation lies in its direct applicability to all professional scenarios without further consideration. High performance on MatCha indicates strong foundational capabilities, but evaluating MLLMs in specific operational laboratory or industrial settings may necessitate adaptation or fine-tuning to those unique professional contexts. Real-world applications often involve distinct instrument variations, proprietary data intricacies, evolving experimental procedures, or specific analytical goals not exhaustively covered by any standardized benchmark. Therefore, there is a risk that strong MatCha scores might be overgeneralized, and deploying these MLLMs effectively into diverse, practical workflows will likely still require careful contextual adjustments and validation to ensure optimal and reliable performance in those specialized environments.

\section*{Acknowledgments}

This work was supported by the Shenzhen Science and Technology Program (JCYJ20220818103001002), Shenzhen Doctoral Startup Funding (RCBS20221008093330065), Tianyuan Fund for Mathematics of National Natural Science Foundation of China (NSFC) (12326608), Shenzhen Science and Technology Program (Shenzhen Key Laboratory Grant No. ZDSYS20230626091302006), and Shenzhen Stability Science Program 2023, Shenzhen Key Lab of Multi-Modal Cognitive Computing. The authors thank the financial support from the National Natural Science Foundation of China (Grant no. 22302174) and the Natural Science Foundation of Zhejiang Province (Grant no. LZ25E030005). Dr. Hong-Qing Liang acknowledges gratefully the research startup package from Zhejiang University.

\bibliography{custom}

\appendix

\section{Sub-task Details}
\label{sec:appendix_subtask}

\begin{table*}[t]
\centering
\resizebox{\textwidth}{!}{
\begin{tabular}{@{}lll@{}}
  \toprule
  \textbf{Stage} & \textbf{Sub-task} & \textbf{Explanation} \\
  \hline
  
  \multirow{2}*{Processing Correlation} & Characterization Technique Identification & Determine the characterization technique \textbf{used} (e.g., SEM, TEM, XRD, AFM). \\
  & Characterization Purpose Inference & Deduce the scientific \textbf{purpose} of using a particular characterization technique. \\
  \hline
  
  \multirow{10}*{Morphology Analysis} & \makecell[l]{Material Classification} & \makecell[l]{Infer the general \textbf{category} of the material (e.g., metal, ceramic, polymer, composite).} \\
  & \makecell[l]{Image Content Analysis \\ $ $} & \makecell[l]{Analyze \textbf{microstructural content from electron microscope images} \\ to extract material characteristics.} \\
  & \makecell[l]{Surface Microstructure Assessment \\ $ $} & \makecell[l]{Determine \textbf{structural features} such as the presence of surface defects, \\ the order of the crystal structure, and the existence of layered structures, etc.} \\
  & \makecell[l]{Surface Roughness Assessment \\ $ $} & \makecell[l]{Evaluate whether the material \textbf{surface} appears smooth or rough, \\ or determine the level of surface roughness.} \\
  & \makecell[l]{Defect Type Classification \\ $ $} & \makecell[l]{Recognize and classify \textbf{defect} types, \\ such as dislocations, vacancies, stacking faults, grain boundaries, etc.} \\
  & \makecell[l]{Grain/Pore Size Classification} & \makecell[l]{Categorize the \textbf{size scale} of grains or pores (e.g., nanometer, micrometer, or millimeter range).} \\
  \hline

  \multirow{16}*{Structure Analysis} & \makecell[l]{Crystallographic Data Inference \\ $ $} & \makecell[l]{Utilize unit \textbf{cell parameters and lattice spacings} to determine structural features, \\ such as symmetry, space group, and overall lattice architecture.} \\
  & \makecell[l]{Crystallinity Classification \\ $ $} & \makecell[l]{Assess whether the material is \textbf{amorphous, polycrystalline, or single crystalline} \\  based on the image.} \\
  & \makecell[l]{Multiphase Interface Assessment \\ $ $} & \makecell[l]{Examine the presence of \textbf{multiple phases or interfaces} within the image, \\ and analyze their structural and compositional features.} \\
  & \makecell[l]{X-ray diffraction (XRD) Pattern Analysis \\ $ $} & \makecell[l]{Extract and analyze key information from \textbf{XRD spectra}, \\ including peak positions and other characteristic features.} \\
  & \makecell[l]{Phase Analysis \\ $ $} & \makecell[l]{Include \textbf{phase identification and classification}, interpretation of phase composition, assess-\\ment of phase homogeneity, determination of crystal structure and polymorphic forms, etc.} \\
  & \makecell[l]{Elemental Mapping Analysis \\ $ $} & \makecell[l]{Identify elements represented by different colors or regions in \textbf{elemental mapping images}, \\ (e.g., EDS or EELS maps).} \\
  & \makecell[l]{Element Distribution Homogeneity Assessment \\ $ $} & \makecell[l]{Analyze the image to assess whether elements are uniformly \textbf{distributed} across the material. \\ $ $} \\
  & \makecell[l]{Material Morphology/Composition Uniformity Assessment \\ $ $} & \makecell[l]{Assess the \textbf{uniformity of material morphology and composition}, such as \\ component ratios, particle size distribution, the homogeneity of internal microstructures, etc.} \\
  \hline

  \multirow{8}*{Property Analysis} & \makecell[l]{Physical and Chemical Properties Inference} & \makecell[l]{Predict potential \textbf{physical or chemical properties} of materials.} \\
  & \makecell[l]{Mechanical Properties Analysis\\ $ $} & \makecell[l]{Extract key parameters such as yield strength, ultimate tensile strength, \\ and ductility from the \textbf{stress-strain curve} to assess material performance.} \\
  & \makecell[l]{Thermal Analysis} & \makecell[l]{Extract critical information from various \textbf{thermal analysis methods} (e.g., TGA, DTA, DSC).} \\
  & \makecell[l]{Infrared (IR) and Raman (RS) Spectral Analysis\\ $ $} & \makecell[l]{Elucidate the \textbf{molecular structure and chemical composition}, identify functional groups, \\ and bond types associated with specific spectral peaks, etc.} \\
  & \makecell[l]{X-ray Photoelectron Spectroscopy (XPS) Spectrum Analysis\\ $ $} & \makecell[l]{Analyze \textbf{XPS spectra} to identify peak positions, determine elemental composition, \\ and chemical states, and infer chemical structures.} \\
  \bottomrule
  
\end{tabular}}
\captionsetup{justification=centering}
\caption{MatCha taxonomy of sub-tasks.}
\label{tab:tasks}
\vspace{-\intextsep}
\end{table*}

We present each stage along with its corresponding sub-tasks and explanation in \cref{tab:tasks}, all defined by materials scientists. These sub-tasks are designed to comprehensively encompass common challenges and characterization techniques encountered in materials characterization.

\section{Dataset Statistics and Diversity}
\label{sec:appendix_data_stats}

Based on our statistics, the papers used in MatCha are sourced from the following journals: Communications Earth \& Environment, Communications Materials, Light: Science \& Applications, NPG Asia Materials, Nature Biotechnology, Nature Communications, Nature Materials, Nature Photonics, Nature Synthesis, Polymer Journal, Scientific Data, Scientific Reports, npj Computational Materials, and npj Heritage Science. The distribution statistics below show that the dataset encompasses a wide range of mainstream characterization techniques and material types. These types are annotated by instructing GPT-4o and are then classified and merged by human experts accordingly. Previous datasets in the materials characterization domain are largely limited to techniques such as SEM and XRD, while MatCha integrates data across the spectrum of mainstream methods, making it the most diverse dataset of its kind to date.

\subsection{Distribution of Characterization Types}
The data in MatCha cover a wide range of characterization techniques, categorized as follows:
\begin{itemize}
    \item \textbf{Microscopy}: Includes transmission electron microscopy (147), scanning electron microscopy (550), scanning transmission electron microscopy (83), optical microscopy (42), X-ray imaging/tomography (1), scanning probe microscopy (11), atom probe tomography (5), focused ion beam-SEM (6), and X-ray photoemission electron microscopy (2).
    \item \textbf{Spectroscopy}: Includes Raman spectroscopy (42), photoluminescence/fluorescence spectroscopy (22), infrared spectroscopy (23), energy-dispersive X-ray spectroscopy (21), X-ray photoelectron spectroscopy (9), nuclear magnetic resonance (10), time-resolved spectroscopy (8), X-ray absorption spectroscopy (6), ultraviolet-visible spectroscopy (5), mass spectrometry (3), extended X-ray absorption fine structure (3), electroluminescence (3), electron probe micro-Analysis (2), Fourier-transform spectroscopy (1), atomic absorption spectroscopy (1), Auger electron spectroscopy (1), electron energy loss spectroscopy (1), electron paramagnetic resonance (1), and cathodoluminescence (1).
    \item \textbf{Diffraction and Scattering}: Includes electron diffraction (68), X-ray diffraction (58), electron backscatter diffraction (18), small-angle X-ray scattering (8), transmission Kikuchi diffraction (3), grazing-incidence X-ray diffraction (2), neutron diffraction (2), and reciprocal space mapping (1).
    \item \textbf{Electrochemical Analysis}: Includes general electrochemical tests (15), cyclic voltammetry (9), electrochemical impedance spectroscopy (3), voltammetry (2), cycling stability test (2), performance evaluation (Faradaic/Coulombic efficiency, 2), galvanostatic intermittent titration technique (1), and galvanostatic charge-discharge (1).
    \item \textbf{Computation and Simulation}: Includes microscopy/diffraction simulation (7), general simulation/computation (7), quantum chemistry calculation (4), first-principles calculation (2), optical simulation (1), phase diagram/thermodynamic calculation (1), finite element method (1), and molecular dynamics (1).
    \item \textbf{Mechanical Testing}: Includes stress-strain test (8), general mechanical testing (5), fracture toughness (1), scratch test (1), and nanoindentation (1).
    \item \textbf{Thermal Analysis}: Includes differential scanning calorimetry (4), thermal imaging (1), and thermal conductivity measurement (1).
    \item \textbf{Magnetic Characterization}: Includes magnetic measurement (3), and X-ray magnetic circular dichroism (2).
    \item \textbf{Other and Performance Evaluation}: Includes data/image analysis (65), electrical/optoelectronic device performance (18), elemental/compositional mapping (10), and physical property measurement (9).
\end{itemize}

\subsection{Distribution of Material Types}
The benchmark covers four major categories of materials: metallic materials (683), inorganic non-metallic materials (435), composite materials (167), and organic polymer materials (98).
\section{Parameters Settings}
\label{sec:appendix_hyp}

To reduce randomness, the temperature is fixed at 0 for models using API interface. For models executed with the Transformers library, the default setting is retained. Furthermore, in zero-shot and few-shot settings where models are required to directly output multiple-choice answers, the maximum generation length (max\_new\_tokens) is set to 32. In contrast, for the CoT experiments, the generation length is set to 8192.
\section{Detailed Few-shot and Chain-of-Thought Results}
\label{sec:appendix_fewshot_cot}

We evaluate model performance under both few-shot and CoT prompting settings. For the few-shot experiments, we randomly sample 2, 4, 8, and 16 in-context examples for each test instance from its corresponding data source. For CoT prompting, we append the phrase “Let’s think step by step” to the end of the original instruction. The results are shown in \cref{tab:main_results_2_shot}, \cref{tab:main_results_4_shot}, \cref{tab:main_results_8_shot}, \cref{tab:main_results_16_shot}, and \cref{tab:main_results_cot}, respectively.

In the 8- and 16-shot settings, several models, including LLaVA-1.5 (7B, 13B) and Janus-Pro-7B, fail to output a valid option, likely due to challenges in processing the extended context.

\begin{table*}[t]
    \resizebox{\textwidth}{!}{
    \begin{tabular}{l|cccccccccc}

    \hline
    \toprule
    \multirow{2}{*}{\textbf{Model}} & \multicolumn{5}{c}{{\cellcolor{lightyellow}\textbf{Generated VQA}}} & \multicolumn{4}{c}{{\cellcolor{lightblue}\textbf{Converted VQA}}} & \multicolumn{1}{c}{\cellcolor{lightred}\textbf{MatCha}}\\

    & \cellcolor{lightyellow}\textbf{PC} & \cellcolor{lightyellow}\textbf{MA} & \cellcolor{lightyellow}\textbf{SA} & \cellcolor{lightyellow}\textbf{PA} & \cellcolor{lightyellow}\textbf{All} & \cellcolor{lightblue}\textbf{Suppl. SMA} & \cellcolor{lightblue}\textbf{Suppl. DTC} & \cellcolor{lightblue}\textbf{Suppl. ICA} & \cellcolor{lightblue}\textbf{All} & \cellcolor{lightred}\textbf{All} \\
    
    \midrule
        \multicolumn{11}{c}{\cellcolor[HTML]{F5F5F5}\textit{Proprietary Models}} \\
    \midrule
    
    GPT-4o \cite{achiam2023gpt}                      & 82.35 & 71.97  & \textbf{71.35} & 67.58  & 62.47 & \textbf{87.64} & 60.61 & 46.12 & 68.97 & 64.67 \\
    GPT-4o-mini \cite{hurst2024gpt}                 & 73.86  & 66.09 & 58.65  & 53.30 & 48.59 & 52.06 & 63.64 & 13.59 & 37.15 & 44.73 \\
    Gemini-1.5-Flash \cite{team2024gemini}            & 76.47 & 63.67 & 58.11 & 58.24 & 51.31 & 63.67 & 60.61 & 33.01 & 50.99 & 51.20 \\
    Gemini-1.5-Pro \cite{team2024gemini}              & 80.39 & 66.78 & 58.92 & 58.24 & 52.72 & 59.18 & 60.61 & 50.97 & 55.93 & 53.80 \\
    Claude-3.5-Sonnet \cite{claude}                 & 74.51 & 70.59 & 67.03 & \textbf{68.68} & 59.66 & 26.59 & \textbf{78.79} & \textbf{57.77} & 42.69 & 53.93 \\
    LlaMA-4-Maverick \cite{meta2025llama}            & \textbf{87.58} & \textbf{75.78} & 70.81 & 67.58 & \textbf{64.59} & 83.52 & 66.67 & 56.80 & \textbf{71.54} & \textbf{66.93} \\

    \midrule
        \multicolumn{11}{c}{\cellcolor[HTML]{F5F5F5}\textit{Open-source Models}} \\
    \midrule
    
    Qwen2.5-VL-7B \cite{qwen2_5}                        & 64.71 & 58.82 & 53.24 & \textbf{59.34} & 43.86 & 38.58 & 57.58 & 24.76 & 34.19 & 40.60 \\
    Qwen2.5-VL-32B \cite{qwen2_5}                       & 64.71 & 65.05 & {60.81} & 57.69 & 48.29 & 39.70 & 51.52 & 24.27 & 34.19 & 43.53 \\
    InternVL3-8B \cite{chen2024expanding}                & 52.29 & 55.71 & 51.89 & 50.00 & 39.74 & 65.92 & 57.58 & \textbf{35.92} & 53.16 & 44.27 \\
    InternVL3-38B \cite{chen2024expanding}               & \textbf{66.67} & \textbf{68.51} & 57.57 & 57.14 & \textbf{50.50} & \textbf{74.91} & \textbf{63.64} & 29.13 & \textbf{55.53} & \textbf{52.20} \\
    LLaVA-1.5-7B \cite{liu2024improved}                 & 26.14 & 35.99 & 34.32 & 31.32 & 23.94 & 32.58 & 33.33 & 12.62 & 24.51 & 24.13 \\
    LLaVA-1.5-13B \cite{liu2024improved}                & 32.68 & 48.44 & 42.70 & 43.41 & 33.00 & 32.21 & 18.18 & 12.62 & 23.32 & 29.73 \\
    Llama-3.2-11B-Vision \cite{grattafiori2024llama}     & 47.06 & 48.10 & 40.00 & 45.05 & 30.99 & 38.95 & 12.12 & 9.71 & 25.30 & 29.07 \\
    Janus-Pro-7B \cite{chen2025janus}                   & 50.33 & 53.98 & 57.57 & 55.49 & 42.66 & 32.21 & 18.18 & 15.05 & 24.31 & 36.47 \\
    Gemma-3-4b-it \cite{team2025gemma}                  & 54.90 & 49.13 & 42.70 & 43.41 & 34.41 & 30.34 & 57.58 & 21.84 & 28.66 & 32.47 \\

    \bottomrule
    \hline
    \end{tabular}
    }
    \caption{2-shot results of model performance on MatCha.
    \textbf{Bolded} values signify the optimal in-class outcomes (open-source or proprietary).
    }
    \label{tab:main_results_2_shot}
    \vspace{-\intextsep}
\end{table*}
\begin{table*}[t]
    \resizebox{\textwidth}{!}{
    \begin{tabular}{l|cccccccccc}

    \hline
    \toprule
    \multirow{2}{*}{\textbf{Model}} & \multicolumn{5}{c}{{\cellcolor{lightyellow}\textbf{Generated VQA}}} & \multicolumn{4}{c}{{\cellcolor{lightblue}\textbf{Converted VQA}}} & \multicolumn{1}{c}{\cellcolor{lightred}\textbf{MatCha}}\\

    & \cellcolor{lightyellow}\textbf{PC} & \cellcolor{lightyellow}\textbf{MA} & \cellcolor{lightyellow}\textbf{SA} & \cellcolor{lightyellow}\textbf{PA} & \cellcolor{lightyellow}\textbf{All} & \cellcolor{lightblue}\textbf{Suppl. SMA} & \cellcolor{lightblue}\textbf{Suppl. DTC} & \cellcolor{lightblue}\textbf{Suppl. ICA} & \cellcolor{lightblue}\textbf{All} & \cellcolor{lightred}\textbf{All} \\
    
    \midrule
        \multicolumn{11}{c}{\cellcolor[HTML]{F5F5F5}\textit{Proprietary Models}} \\
    \midrule
    
    GPT-4o \cite{achiam2023gpt}                      & 84.31 & 73.01 & \textbf{74.05} & 65.93 & 63.88 & 85.02 & 57.58 & 53.88 & 70.55 & 66.13 \\
    GPT-4o-mini \cite{hurst2024gpt}                 & 77.78 & 66.44 & 60.27 & 55.49 & 51.11 & 47.57 & 54.55 & 16.50 & 35.38 & 45.80 \\
    Gemini-1.5-Flash \cite{team2024gemini}            & 78.43 & 62.63 & 57.84 & 56.04 & 50.30 & 35.21 & 57.58 & 33.01 & 35.77 & 45.40 \\
    Gemini-1.5-Pro \cite{team2024gemini}              & 83.01 & 70.59 & 65.41 & 63.19 & 58.15 & 59.55 & 66.67 & 47.09 & 54.94 & 57.07 \\
    Claude-3.5-Sonnet \cite{claude}                 & 87.58 & \textbf{74.39} & 69.46 & \textbf{69.23} & \textbf{64.59} & 81.65 & \textbf{75.76} & \textbf{68.93} & \textbf{76.09} & \textbf{68.47} \\
    LlaMA-4-Maverick \cite{meta2025llama}            & \textbf{88.24} & 71.28 & 72.70 & 67.03 & 63.98 & \textbf{85.39} & 66.67 & 50.00 & 69.76 & 65.93 \\

    \midrule
        \multicolumn{11}{c}{\cellcolor[HTML]{F5F5F5}\textit{Open-source Models}} \\
    \midrule
    
    Qwen2.5-VL-7B \cite{qwen2_5}                        & 62.75 & 62.63 & 51.35 & \textbf{58.79} & 43.76 & 46.82 & 51.52 & 20.87 & 36.56 & 41.33 \\
    Qwen2.5-VL-32B \cite{qwen2_5}                       & 63.40 & 67.13 & \textbf{59.46} & 57.14 & 48.79 & 48.69 & 48.48 & 23.30 & 38.34 & 45.27 \\
    InternVL3-8B \cite{chen2024expanding}                & 50.33 & 56.75 & 51.89 & 49.45 & 41.35 & 50.19 & 51.52 & 33.98 & 43.68 & 42.13 \\
    InternVL3-38B \cite{chen2024expanding}               & \textbf{70.59} & \textbf{69.90} & 58.92 & \textbf{58.79} & \textbf{51.41} & \textbf{60.67} & \textbf{57.58} & \textbf{40.78} & \textbf{52.37} & \textbf{51.73} \\
    LLaVA-1.5-7B \cite{liu2024improved}                & 21.57 & 23.88 & 31.62 & 28.02 & 17.81 & 32.21 & 21.21 & 13.59 & 23.91 & 19.87 \\
    LLaVA-1.5-13B \cite{liu2024improved}               & 32.03 & 24.22 & 30.54 & 25.27 & 18.91 & 32.21 & 24.24 & 13.59 & 24.11 & 20.67 \\
    Llama-3.2-11B-Vision \cite{grattafiori2024llama}    & 47.06 & 54.67 & 44.32 & 45.60 & 35.21 & 23.97 & 18.18 & 16.99 & 20.75 & 30.33 \\
    Janus-Pro-7B \cite{chen2025janus}                   & 52.29 & 54.33 & 52.43 & 51.10 & 39.64 & 33.33 & 27.27 & 15.05 & 25.49 & 34.87 \\
    Gemma-3-4b-it \cite{team2025gemma}                  & 50.98 & 47.75 & 44.32 & 43.96 & 32.80 & 33.33 & 42.42 & 24.76 & 30.43 & 32.00 \\

    \bottomrule
    \hline
    \end{tabular}
    }
    \caption{4-shot results of model performance on MatCha.
    \textbf{Bolded} values signify the optimal in-class outcomes (open-source or proprietary).
    }
    \label{tab:main_results_4_shot}
    \vspace{-\intextsep}
\end{table*}
\begin{table*}[t]
    \resizebox{\textwidth}{!}{
    \begin{tabular}{l|cccccccccc}

    \hline
    \toprule
    \multirow{2}{*}{\textbf{Model}} & \multicolumn{5}{c}{{\cellcolor{lightyellow}\textbf{Generated VQA}}} & \multicolumn{4}{c}{{\cellcolor{lightblue}\textbf{Converted VQA}}} & \multicolumn{1}{c}{\cellcolor{lightred}\textbf{MatCha}}\\

    & \cellcolor{lightyellow}\textbf{PC} & \cellcolor{lightyellow}\textbf{MA} & \cellcolor{lightyellow}\textbf{SA} & \cellcolor{lightyellow}\textbf{PA} & \cellcolor{lightyellow}\textbf{All} & \cellcolor{lightblue}\textbf{Suppl. SMA} & \cellcolor{lightblue}\textbf{Suppl. DTC} & \cellcolor{lightblue}\textbf{Suppl. ICA} & \cellcolor{lightblue}\textbf{All} & \cellcolor{lightred}\textbf{All} \\
    
    \midrule
        \multicolumn{11}{c}{\cellcolor[HTML]{F5F5F5}\textit{Proprietary Models}} \\
    \midrule
    
    GPT-4o \cite{achiam2023gpt}                      & 84.97 & 73.36 & \textbf{72.70} & 68.13 & \textbf{64.19} & 86.52 & 60.61 & 46.12 & 68.38 & 65.60 \\
    GPT-4o-mini \cite{hurst2024gpt}                 & 77.12 & 63.32 & 59.73 & 57.14 & 49.70 & 63.67 & 54.55 & 18.93 & 44.86 & 48.07 \\
    Gemini-1.5-Flash \cite{team2024gemini}            & 81.05 & 64.71 & 57.03 & 57.14 & 50.70 & 52.81 & 33.33 & 43.20 & 47.63 & 49.67 \\
    Gemini-1.5-Pro \cite{team2024gemini}              & 84.31 & 69.20 & 60.00 & 66.48 & 56.94 & 64.42 & 66.67 & 43.69 & 56.13 & 56.67 \\
    Claude-3.5-Sonnet \cite{claude}                 & \textbf{85.62} & \textbf{75.43} & 67.84 & \textbf{69.23} & 63.68 & 89.89 & \textbf{84.85} & \textbf{62.14} & \textbf{78.26} & \textbf{68.60} \\
    LlaMA-4-Maverick \cite{meta2025llama}            & 84.31 & 73.36 & 71.89 & 66.48 & 63.38 & \textbf{90.26} & 60.61 & 52.43 & 72.92 & 66.60 \\

    \midrule
        \multicolumn{11}{c}{\cellcolor[HTML]{F5F5F5}\textit{Open-source Models}} \\
    \midrule
    
    Qwen2.5-VL-7B \cite{qwen2_5}                        & 71.24 & 61.94 & 53.78 & 57.14 & 45.47 & 32.96 & 48.48 & 20.87 & 29.05 & 39.93 \\
    Qwen2.5-VL-32B \cite{qwen2_5}                       & 62.09 & 67.47 & \textbf{60.27} & \textbf{60.44} & 50.00 & 38.58 & \textbf{57.58} & 20.39 & 32.41 & 44.07 \\
    InternVL3-8B \cite{chen2024expanding}                & 56.86 & 57.44 & 52.16 & 50.55 & 42.05 & \textbf{60.30} & 48.48 & \textbf{35.92} & \textbf{49.60} & 44.60 \\
    InternVL3-38B \cite{chen2024expanding}               & \textbf{73.20} & \textbf{67.47} & 59.46 & 55.49 & \textbf{51.51} & 58.43 & 54.55 & 33.98 & 48.22 & \textbf{50.40} \\
    LLaVA-1.5-7B \cite{liu2024improved}                & 0.00 & 0.00 & 0.00 & 0.00 & 0.00 & 0.00 & 0.00 & 0.00 & 0.00 & 0.00 \\
    LLaVA-1.5-13B \cite{liu2024improved}               & 0.00 & 0.00 & 0.00 & 0.00 & 0.00 & 0.00 & 0.00 & 0.00 & 0.00 & 0.00 \\
    Llama-3.2-11B-Vision \cite{grattafiori2024llama}    & 51.63 & 50.87 & 45.14 & 42.31 & 32.49 & 36.70 & 9.09 & 16.02 & 26.48 & 30.47 \\
    Janus-Pro-7B \cite{chen2025janus}                   & 0.00 & 0.00 & 0.00 & 0.00 & 0.00 & 0.00 & 0.00 & 0.00 & 0.00 & 0.00 \\
    Gemma-3-4b-it \cite{team2025gemma}                  & 58.17 & 49.48 & 47.03 & 46.70 & 37.02 & 36.33 & 39.39 & 23.79 & 31.42 & 35.13 \\

    \bottomrule
    \hline
    \end{tabular}
    }
    \caption{8-shot results of model performance on MatCha.
    \textbf{Bolded} values signify the optimal in-class outcomes (open-source or proprietary).
    }
    \label{tab:main_results_8_shot}
    \vspace{-\intextsep}
\end{table*}
\begin{table*}[t]
    \resizebox{\textwidth}{!}{
    \begin{tabular}{l|cccccccccc}

    \hline
    \toprule
    \multirow{2}{*}{\textbf{Model}} & \multicolumn{5}{c}{{\cellcolor{lightyellow}\textbf{Generated VQA}}} & \multicolumn{4}{c}{{\cellcolor{lightblue}\textbf{Converted VQA}}} & \multicolumn{1}{c}{\cellcolor{lightred}\textbf{MatCha}}\\

    & \cellcolor{lightyellow}\textbf{PC} & \cellcolor{lightyellow}\textbf{MA} & \cellcolor{lightyellow}\textbf{SA} & \cellcolor{lightyellow}\textbf{PA} & \cellcolor{lightyellow}\textbf{All} & \cellcolor{lightblue}\textbf{Suppl. SMA} & \cellcolor{lightblue}\textbf{Suppl. DTC} & \cellcolor{lightblue}\textbf{Suppl. ICA} & \cellcolor{lightblue}\textbf{All} & \cellcolor{lightred}\textbf{All} \\
    
    \midrule
        \multicolumn{11}{c}{\cellcolor[HTML]{F5F5F5}\textit{Proprietary Models}} \\
    \midrule
    
    GPT-4o \cite{achiam2023gpt}                      & 83.66 & 73.36 & \textbf{73.51} & \textbf{73.08} & \textbf{65.19} & 89.14 & 66.67 & 54.37 & 73.52 & 68.00 \\
    GPT-4o-mini \cite{hurst2024gpt}                 & 78.43 & 68.17 & 60.81 & 57.14 & 52.62 & 64.79 & 60.61 & 16.99 & 45.06 & 50.07 \\
    Gemini-1.5-Flash \cite{team2024gemini}            & 77.12 & 63.67 & 57.84 & 58.24 & 51.21 & 61.80 & 57.58 & 44.66 & 54.55 & 52.33 \\
    Gemini-1.5-Pro \cite{team2024gemini}              & \textbf{84.97} & 65.40 & 60.54 & 61.54 & 55.53 & 59.18 & 66.67 & 34.47 & 49.60 & 53.53 \\
    Claude-3.5-Sonnet \cite{claude}                 & 83.01 & 73.70 & 67.57 & 68.13 & 62.07 & 75.28 & \textbf{75.76} & 41.26 & 61.46 & 61.87 \\
    LlaMA-4-Maverick \cite{meta2025llama}            & \textbf{84.97} & \textbf{75.09} & 71.08 & 68.68 & 64.29 & \textbf{89.89} & 42.42 & \textbf{62.62} & \textbf{75.69} & \textbf{68.13} \\

    \midrule
        \multicolumn{11}{c}{\cellcolor[HTML]{F5F5F5}\textit{Open-source Models}} \\
    \midrule
    
    Qwen2.5-VL-7B \cite{qwen2_5}                        & 70.59 & 62.63 & 54.05 & 58.24 & 46.08 & 33.33 & 48.48 & 10.19 & 24.90 & 38.93 \\
    Qwen2.5-VL-32B \cite{qwen2_5}                       & 64.05 & 66.09 & \textbf{62.97} & \textbf{60.44} & 50.50 & 40.07 & 51.52 & 24.76 & 34.58 & 45.13 \\
    InternVL3-8B \cite{chen2024expanding}                & 62.09 & 61.94 & 52.97 & 53.85 & 45.37 & 47.94 & 45.45 & 33.01 & 41.70 & 44.13 \\
    InternVL3-38B \cite{chen2024expanding}               & \textbf{75.16} & \textbf{69.55} & 61.08 & 59.34 & \textbf{53.22} & \textbf{49.81} & \textbf{54.55} & \textbf{36.41} & \textbf{44.66} & \textbf{50.33} \\
    LLaVA-1.5-7B \cite{liu2024improved}                & 0.00 & 0.00 & 0.00 & 0.00 & 0.00 & 0.00 & 0.00 & 0.00 & 0.00 & 0.00 \\
    LLaVA-1.5-13B \cite{liu2024improved}               & 0.00 & 0.00 & 0.00 & 0.00 & 0.00 & 0.00 & 0.00 & 0.00 & 0.00 & 0.00 \\
    Llama-3.2-11B-Vision \cite{grattafiori2024llama}    & 57.52 & 58.13 & 45.68 & 40.11 & 37.12 & 36.33 & 18.18 & 10.68 & 24.70 & 32.93 \\
    Janus-Pro-7B \cite{chen2025janus}                   & 55.56 & 55.36 & 61.35 & 57.14 & 46.38 & 31.09 & 45.45 & 21.36 & 28.06 & 40.20 \\
    Gemma-3-4b-it \cite{team2025gemma}                  & 64.71 & 50.87 & 49.46 & 42.86 & 38.53 & 28.09 & 39.39 & 24.76 & 27.47 & 34.80 \\

    \bottomrule
    \hline
    \end{tabular}
    }
    \caption{16-shot results of model performance on MatCha.
    \textbf{Bolded} values signify the optimal in-class outcomes (open-source or proprietary).
    }
    \label{tab:main_results_16_shot}
    \vspace{-\intextsep}
\end{table*}

\begin{table*}[t]
    \resizebox{\textwidth}{!}{
    \begin{tabular}{l|cccccccccc}

    \hline
    \toprule
    \multirow{2}{*}{\textbf{Model}} & \multicolumn{5}{c}{{\cellcolor{lightyellow}\textbf{Generated VQA}}} & \multicolumn{4}{c}{{\cellcolor{lightblue}\textbf{Converted VQA}}} & \multicolumn{1}{c}{\cellcolor{lightred}\textbf{MatCha}}\\

    & \cellcolor{lightyellow}\textbf{PC} & \cellcolor{lightyellow}\textbf{MA} & \cellcolor{lightyellow}\textbf{SA} & \cellcolor{lightyellow}\textbf{PA} & \cellcolor{lightyellow}\textbf{All} & \cellcolor{lightblue}\textbf{Suppl. SMA} & \cellcolor{lightblue}\textbf{Suppl. DTC} & \cellcolor{lightblue}\textbf{Suppl. ICA} & \cellcolor{lightblue}\textbf{All} & \cellcolor{lightred}\textbf{All} \\
    
    \midrule
        \multicolumn{11}{c}{\cellcolor[HTML]{F5F5F5}\textit{Proprietary Models}} \\
    \midrule
    
    GPT-4o \cite{achiam2023gpt}                      & \textbf{84.97} & \textbf{73.36}  & \textbf{68.65} & \textbf{70.88}  & \textbf{63.08} & {55.43} & 69.70 & 46.12 & 52.57 & \textbf{59.53} \\
    GPT-4o-mini \cite{hurst2024gpt}                 & 71.90  & 57.09 & 61.35  & 58.24 & 48.39 & 53.93 & 54.55 & 24.76 & 42.09 & 46.27 \\
    Gemini-1.5-Flash \cite{team2024gemini}            & 78.43 & 61.94 & 56.49 & 59.89 & 49.70 & 47.19 & 42.42 & 30.58 & 40.12 & 46.47 \\
    Gemini-1.5-Pro \cite{team2024gemini}              & 83.66 & 69.55 & 60.81 & 64.29 & 55.63 & 47.19 & 72.73 & 42.23 & 46.84 & 52.67 \\
    Claude-3.5-Sonnet \cite{claude}                 & 84.31 & 71.97 & 68.11 & 69.78 & 61.77 & 49.06 & \textbf{81.82} & 51.46 & 52.17 & 58.53 \\
    LlaMA-4-Maverick \cite{meta2025llama}            & 60.78 & 54.33 & 51.62 & 54.95 & 40.85 & 55.06 & 54.55 & \textbf{55.34} & \textbf{55.14} & 45.67 \\

    \midrule
        \multicolumn{11}{c}{\cellcolor[HTML]{F5F5F5}\textit{Open-source Models}} \\
    \midrule
    
    Qwen2.5-VL-7B \cite{qwen2_5}                        & 62.75 & 56.40 & 50.81 & 57.69 & 44.57 & 49.06 & \textbf{60.61} & 22.82 & 39.13 & 42.73 \\
    Qwen2.5-VL-32B \cite{qwen2_5}                       & \textbf{67.32} & \textbf{67.82} & 61.08 & \textbf{64.29} & \textbf{52.52} & 53.18 & \textbf{60.61} & 32.52 & 45.26 & 50.07 \\
    InternVL3-8B \cite{chen2024expanding}                & 45.10 & 54.67 & 52.16 & 56.04 & 40.44 & 53.93 & 51.52 & 29.61 & 43.87 & 41.60 \\
    InternVL3-38B \cite{chen2024expanding}               & 60.78 & 65.40 & 60.27 & 61.54 & 50.30 & \textbf{59.18} & 57.58 & \textbf{41.75} & \textbf{51.98} & \textbf{50.87} \\
    LLaVA-1.5-7B \cite{liu2024improved}                & 5.23 & 15.57 & 15.95 & 22.53 & 9.26 & 32.58 & 24.24 & 10.68 & 23.12 & 13.93 \\
    LLaVA-1.5-13B \cite{liu2024improved}               & 5.88 & 16.96 & 16.49 & 21.98 & 9.46 & 35.58 & 24.24 & 15.05 & 26.48 & 15.20 \\
    Llama-3.2-11B-Vision \cite{grattafiori2024llama}        & 60.13 & 45.67 & 47.03 & 46.70 & 37.32 & 35.96 & 24.24 & 22.82 & 29.84 & 34.80 \\
    Janus-Pro-7B \cite{chen2025janus}                   & 55.56 & 55.71 & \textbf{61.35} & 57.69 & 46.48 & 31.09 & 45.45 & 21.36 & 28.06 & 40.27 \\
    Gemma-3-4b-it \cite{team2025gemma}                  & 58.17 & 49.13 & 49.46 & 45.05 & 36.82 & 28.84 & 57.58 & 24.27 & 28.85 & 34.13 \\

    \bottomrule
    \hline
    \end{tabular}
    }
    \caption{CoT results of model performance on MatCha.
    \textbf{Bolded} values signify the optimal in-class outcomes (open-source or proprietary).
    }
    \label{tab:main_results_cot}
    \vspace{-\intextsep}
\end{table*}

\section{Ablations Experiments}
\label{sec:appendix_abla}

Followed by previous research \cite{goyal2017making, chen2024we}, we conduct a no-image ablation study to investigate the contribution of visual information, shown in \cref{tab:no_image}.

The notably low score of LLaMA-4-Maverick in the no-image condition is primarily due to its tendency to bypass the question and directly generate answers without adequately adhering to instructions in the absence of visual input. Other models also exhibited substantial performance degradation--over 20\%--with accuracy only marginally (around 10\%) above random guessing. This underscores the significant role of visual information in MatCha and highlights the rigorous demands on the visual understanding capabilities of models.

\begin{table}[H]
    \centering
    
    \resizebox{\columnwidth}{!}{
        \begin{tabular}{l|cc}
        \hline
        \toprule
        \multirow{2}{*}{\textbf{Model}} & MatCha & Ablation \\
        & All & no-image drop \\
        \midrule
        GPT-4o \cite{achiam2023gpt} & 34.33 & -24.74 \\
        Gemini-1.5-Pro \cite{team2024gemini} & 29.53 & -26.87 \\
        Claude-3.5-Sonnet \cite{claude} & 34.60 & -23.40 \\
        LlaMA-4-Maverick \cite{meta2025llama} & 5.93 & -20.47 \\
        \bottomrule
        \end{tabular}
    }
    \caption{No-image ablation study on MatCha.}
    \label{tab:no_image}
\end{table}

\section{Future Direction: The Potential Role of Retrieval-Augmented Generation}
\label{sec:appendix_rag}

A primary conclusion from our benchmark is that MLLMs are significantly constrained by a lack of specialized domain knowledge, a common challenge in vertical domains. A promising solution is Retrieval-Augmented Generation (RAG), which allows models to access external knowledge bases. RAG has already shown success in scientific domains, such as PaperQA \cite{lala2023paperqa}, which answers complex questions by retrieving information from scientific articles. 

Our few-shot experiments indicate that providing in-context examples can, to some extent, elicit domain knowledge of a model. Similarly, RAG is another powerful paradigm providing dynamic, external knowledge to a model before it generates a response. Given that both methods function by supplying context, we believe RAG can also further improve performance in materials characterization scenarios. This makes MatCha an ideal testbed for evaluating future multimodal RAG systems in materials science.
\clearpage
\onecolumn

\section{Cases}
\label{sec:appendix_cases}

\subsection{VQA Cases}
\label{sec:appendix_vqa_cases}

\begin{AIbox}[breakable]{\textit{Physical and Chemical Properties Inference}}
\vspace{2mm}
\begin{wrapfigure}{R}{0.5\linewidth}
    \centering
    \includegraphics[width=0.85\linewidth]{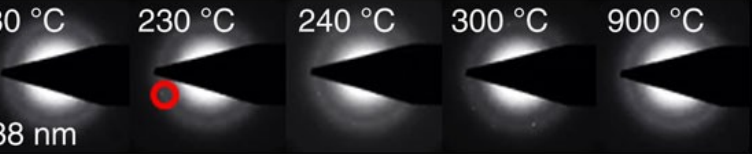}
\end{wrapfigure}

\textbf{Question}: What does the red circle in the 230 °C frame indicate regarding the nanorods' crystallization?

\textbf{Choices}: (A) The maximum diffraction intensity (B) Onset of the first diffraction spot (C) Completion of crystallization (D) Absence of any crystallization

\textbf{Answer}: B

\vspace{2mm}
    
\end{AIbox}

\begin{AIbox}[breakable]{\textit{Crystallographic Data Inference}}
\vspace{4.5mm}
\begin{wrapfigure}{R}{0.5\linewidth}
    \centering
    \vspace{-1.15cm}
    \includegraphics[width=0.6\linewidth]{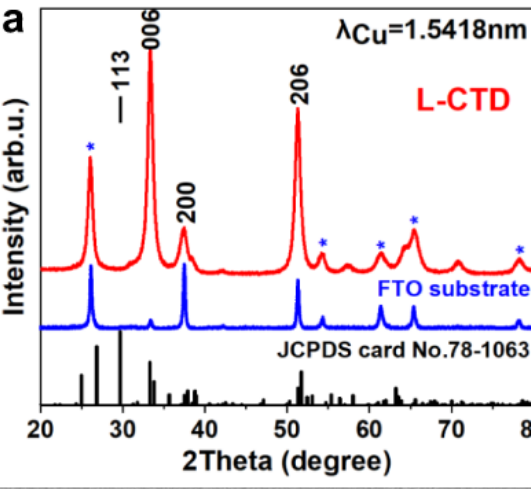}
\end{wrapfigure}

\textbf{Question}: What is the space group designation associated with the phase identified in the L-CTD XRD pattern?

\textbf{Choices}: (A) Fm-3m (B) Pbcn (C) P63/mmc (D) Ia3d

\textbf{Answer}: B

\vspace{5.5mm}

\end{AIbox}

\begin{AIbox}[breakable]{\textit{Surface Microstructure Assessment}}
\vspace{4mm}
\begin{wrapfigure}{R}{0.5\linewidth}
    \centering
    \vspace{-0.8cm}
    \includegraphics[height=0.55\linewidth]{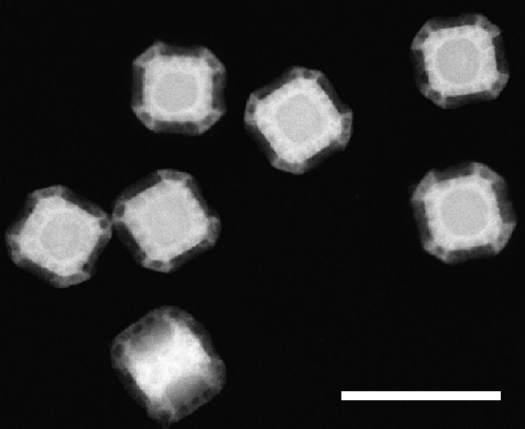}
\end{wrapfigure}

\textbf{Question}: What structural feature is clearly visible in the HAADF-STEM image of type 2 p-ANHs?

\textbf{Choices}: (A) Cylindrical canyons on the {100} facets (B) Entirely smooth surface (C) Spherical nanoparticle morphology (D) Randomly structured pores

\textbf{Answer}: A

\vspace{4mm}
    
\end{AIbox}

\begin{AIbox}[breakable]{\textit{Material Morphology and Composition Uniformity Assessment}}
\vspace{4mm}
\begin{wrapfigure}{R}{0.5\linewidth}
    \centering
    \vspace{-0.8cm}
    \includegraphics[height=0.55\linewidth]{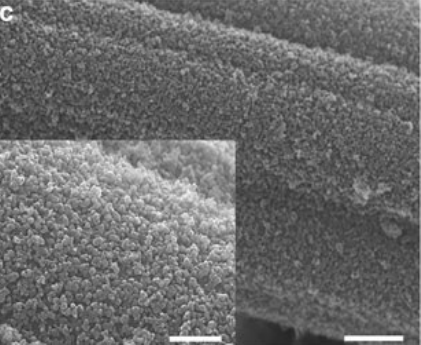}
\end{wrapfigure}

\textbf{Question}: What type of material morphology is observed in the SEM image?

\textbf{Choices}: (A) Rod-like structures (B) Uniformly distributed spherical nanoparticles (C) Thin film coating (D) Porous sponge-like structures

\textbf{Answer}: B

\vspace{5mm}
    
\end{AIbox}

\begin{AIbox}[breakable]{\textit{Mechanical Properties Analysis}}
\vspace{2mm}
\begin{wrapfigure}{R}{0.5\linewidth}
    \centering
    \vspace{-0.6cm}
    \includegraphics[height=0.7\linewidth]{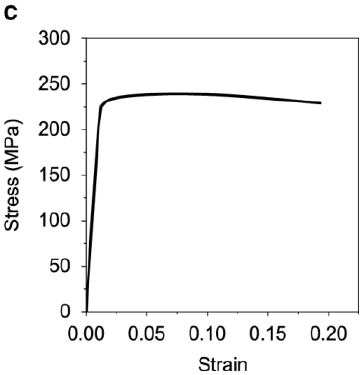}
\end{wrapfigure}

\textbf{Question}: What is the approximate value of the yield stress observed in the engineering stress-strain curve?

\textbf{Choices}: (A) 230 MPa (B) 300 MPa (C) 200 MPa (D) 50 MPa

\textbf{Answer}: A

\vspace{1mm}

\noindent\rule{\linewidth}{0.4pt}

\vspace{1mm}

\textbf{Question}: What is the approximate strain at which stress begins to decrease in the engineering stress-strain curve?

\textbf{Choices}: (A) 0.08 (B) 0.20 (C) 0.01 (D) 0.50

\textbf{Answer}: A
    
\end{AIbox}

\begin{AIbox}[breakable]{\textit{Characterization Technique Identification}}
\vspace{4mm}
\begin{wrapfigure}{R}{0.5\linewidth}
    \centering
    \vspace{-0.85cm}
    \includegraphics[width=0.9\linewidth]{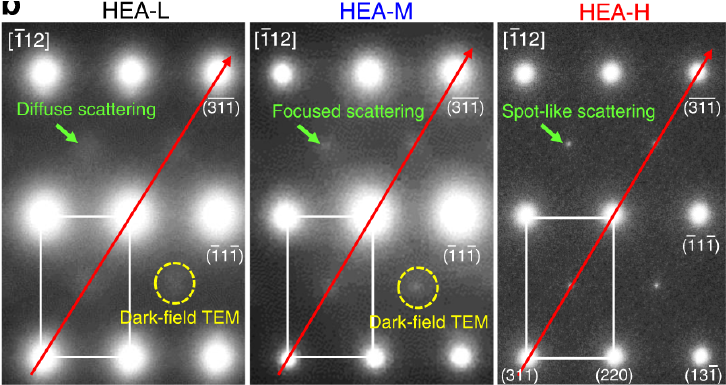}
\end{wrapfigure}

\textbf{Question}: What characterization technique was used to capture the diffraction patterns shown in Figure 5b?

\textbf{Choices}: (A) SEM (B) XRD (C) TEM (D) STEM

\textbf{Answer}: C

\vspace{4mm}
    
\end{AIbox}

\subsection{Error Cases}
\label{sec:appendix_error_cases}

\begin{AIbox}{\textit{Lack of Material Knowledge}}

\begin{wrapfigure}{R}{0.5\linewidth}
    \centering
    \vspace{-0.6cm}
    \includegraphics[width=0.6\linewidth]{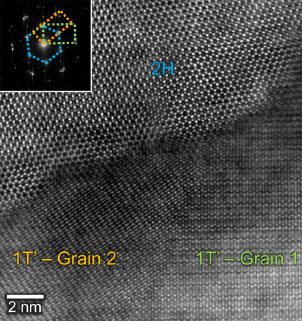}
\end{wrapfigure}

\textbf{Question}: How many distinct lattice structures are identified from the FFT pattern inset? (A) One (B) Two (C) Three (D) Four

\textbf{Correct Answer}: (C)

\textbf{Model Answer}: (B)

\textbf{Model Prediction Rationable}:

To determine how many distinct lattice structures are identified from the FFT (Fast Fourier Transform) pattern inset, let's analyze the image step by step:

1. Observation of the FFT Pattern: 

   The inset FFT pattern shows three distinct sets of diffraction spots, each marked with different colored dashed lines (blue, orange, and green). These correspond to different lattice structures or orientations.

2. Identification of Lattice Structures in the Image:  

   - The main image shows three regions labeled as "2H," "1T' – Grain 1," and "1T' – Grain 2."
   
   - The "2H" region corresponds to one lattice structure.
   
   - The "1T' – Grain 1" and "1T' – Grain 2" regions correspond to the same lattice structure (1T') but with different grain orientations.

3. Conclusion: 

   From the FFT pattern, we can identify two distinct lattice structures:  
   
   - The "2H" lattice structure.  
   
   - The "1T'" lattice structure (with two grains but the same structure).

Thus, the correct answer is: (B) Two

\textbf{Error Analysis}:

The model incorrectly concluded that there are only two distinct lattice structures based on the FFT pattern and the labeled regions in the image. The mistake lies in the interpretation of the FFT pattern inset, which clearly shows three distinct sets of diffraction spots marked by different colored dashed lines (blue, orange, and green). These correspond to three distinct lattice structures or orientations. The model failed to recognize that the FFT pattern indicates three distinct lattice structures, not just two, despite the labeling of '1T' grains in the main image. This error stems from a lack of material knowledge and misunderstanding of how FFT patterns correspond to lattice structures.

\end{AIbox}

\begin{AIbox}{\textit{Visual Perception Error}}

\begin{wrapfigure}{R}{0.5\linewidth}
    \centering
    \vspace{-0.6cm}
    \includegraphics[width=0.6\linewidth]{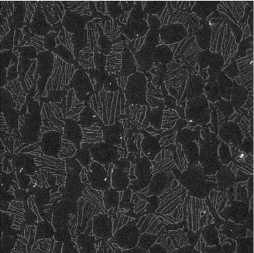}
\end{wrapfigure}

\textbf{Question}: What are the main microstructure in the given scanning electron microscope (SEM) images of Ti-6Al-4V alloy? (A) acicular: acicular morphology (B) lamellar: $\alpha$ lamellae (C) duplex: duplex microstructure, a bimodal morphology consisting of equiaxed grains and $\alpha$-lamellae

\textbf{Correct Answer}: (C)

\textbf{Model Answer}: (B)

\textbf{Model Prediction Rationable}:

To identify the main microstructure in the given SEM image of the Ti-6Al-4V alloy, let's analyze the features step by step:

1. Observation of the Image:

   - The image shows a clear pattern of elongated, parallel structures within larger grains.
   
   - These structures resemble lamellae, which are characteristic of a lamellar microstructure.

2. Comparison with Descriptions:

   - Acicular: Acicular morphology typically consists of needle-like structures, which are not observed in this image.
   
   - Lamellar: The image shows $\alpha$ lamellae within the grains, matching the description of a lamellar microstructure.
   
   - Duplex: A duplex microstructure would show a combination of equiaxed grains and $\alpha$ lamellae. However, the image does not show distinct equiaxed grains alongside lamellae.

3. Conclusion:
   - The microstructure in the SEM image corresponds to (B) lamellar: $\alpha$ lamellae.

\textbf{Error Analysis}:

The model incorrectly identified the microstructure as lamellar (B) instead of duplex (C). The SEM image clearly shows a bimodal morphology consisting of equiaxed grains and $\alpha$-lamellae, which is characteristic of a duplex microstructure. The model failed to recognize the presence of equiaxed grains in the image, focusing only on the lamellae structures. This indicates a visual perception error, as the model did not correctly identify the equiaxed grains in the image.

\end{AIbox}
\section{Prompt Setting}
\label{sec:appendix_prompt}

\subsection{Sub-caption Segmentation Prompt}
\label{sec:appd_prompt_1}

\begin{formattedprompt}
System prompt:\\

Subfigure labels are letters referring to individual subfigures within a larger figure. 
Please separate the given full caption into the exact subcaptions and format as a syntactically valid JSON format with keys the letter of each subcaption. 
If there is no full caption then return an empty JSON.\\

User prompt:\\

Caption:
\{caption\}
\end{formattedprompt}

\subsection{QA Generation Prompt}
\label{sec:appd_prompt_2}

\begin{formattedprompt}
System prompt:\\

You are a scientific expert. Based on the following figure-caption pair and its related context from the article in the field of materials science and characterization, generate 1 to 4 visual question answering (VQA)-style question–answer pairs, depending on the amount of information provided.
The question should be related to both the figure-caption pair and the context, but the answer should be able to be inferred and analyzed ONLY from the figure.

\{sub-tasks with explanation\}

You should generate multiple-choice questions, ensure that the answers are concise and clear. For multiple-choice questions, please generate plausible but incorrect options. The number of options is not limited, and enclose all options in parentheses (e.g., (A)) as part of the question.
After providing the question and answer, also include the topic of this question, and output in JSON format:
\{
  "vqas": [
    {{
      "question": ...,
      "answer": ...,
      "topic": ...
    }}
  ]
\}
Example template:
\{
  "vqas": [
    {{
      "question": ... (A) xx (B) xx (C) xx (D) xx,
      "answer": "B",
      "topic": ...
    }}
  ]
\}\\

User prompt:\\

Figure \{sub-figure\} \\
Caption: \{sub-caption\} \\
Related Context: \{related context\}
\end{formattedprompt}

\subsection{Evaluation Prompt}
\label{sec:appd_prompt_3}

\begin{formattedprompt}
System prompt:\\

You are a helpful materials science assistant. Based on the figure, please answer the following question. The answer could be inferred from the figure and must be concise and clear. Answer directly without any explanation.\\

User prompt:\\

Question: \{question with options\}\\
Answer with the option's letter from the given choices directly:
\end{formattedprompt}

\end{document}